\documentclass[10pt,journal,compsoc]{IEEEtran}

\ifCLASSINFOpdf
 \usepackage[pdftex]{graphicx}
\else
\usepackage[dvips]{graphicx}
\DeclareGraphicsExtensions{.eps}
\fi
\usepackage{amsmath}
\usepackage{amssymb}
\usepackage{amsthm}

\usepackage{amsfonts}
\usepackage{algorithmic}
\usepackage{algorithm}
\usepackage{array}
\usepackage{textcomp}
\usepackage{stfloats}
\usepackage{url}
\usepackage{verbatim}
\usepackage{pifont,booktabs,amssymb,hyperref}
\usepackage{natbib}
\setcitestyle{numbers,square}
\usepackage{nicefrac}       % compact symbols for 1/2, etc.
\usepackage{microtype}      % microtypography
\usepackage{xcolor}         % colors
\usepackage{multirow}
\usepackage{amsmath}
\usepackage{xcolor}         % colors
\usepackage{booktabs} % 用于更优雅的表格线
\usepackage{multirow} %合并单元格
\usepackage{makecell}
\usepackage{pifont} %写带圈数字用
\usepackage{circledsteps}
\usepackage{makecell}%用与生成跨行单元格
\usepackage{tabularx}%定制表宽度
\usepackage[accsupp]{axessibility}  
\usepackage{graphicx}
\usepackage{amssymb}
\usepackage{enumitem}
\usepackage{subcaption}
\usepackage{bbding}
\newcommand{\myMethodName}{TransMamba}

\newcommand{\eg}{\emph{e.g.}{}}

\newcommand{\etc}{\emph{etc}{}}
\newcommand{\cf}{\emph{cf. }{}}

\definecolor{crosscolor}{rgb}{0.969,0.580,0.114} %
\definecolor{checkcolor}{rgb}{0.485,0.640,0.204} %
\def\cmark{{\color{checkcolor}\ding{52}}}
\def\xmark{{\color{crosscolor}\ding{56}}}

\begin{document}

\title{TransMamba: Fast Universal Architecture Adaption from Transformers to Mamba}

\author{Xiuwei~Chen, Wentao Hu, Xiao Dong, Sihao Lin, Zisheng Chen, \\ Meng Cao, Yina Zhuang, Jianhua Han, Hang Xu, Xiaodan Liang\IEEEauthorrefmark{2}
\IEEEcompsocitemizethanks{
\IEEEcompsocthanksitem 
\IEEEauthorrefmark{2}Xiaodan Liang is the corresponding author.\protect\\
\IEEEcompsocthanksitem Xiuwei Chen, Zisheng Chen and Yina Zhuang is with Shenzhen Campus of Sun Yat-sen University, Shenzhen, China.   \protect\\
E-mail: \{chenxw83@mail2.sysu.edu.cn,  halveschen@163.com, zhuangyn3@mail2.sysu.edu.cn\}
\IEEEcompsocthanksitem Wentao Hu is with Hong Kong Polytechnic University, Hongkong, China. \protect\\
 E-mail: wayne-wt.hu@connect.polyu.hk.
\IEEEcompsocthanksitem %Bingqian Lin, 
    Xiao Dong is with Zhuhai Campus of Sun Yat-sen University, Zhuhai, China. \protect\\
			E-mail:\{dongx55@mail2.sysu.edu.cn,\}
\IEEEcompsocthanksitem %Bingqian Lin, 
    Sihao Lin is a postdoc researcher now with University of Adelaide, South Australia. \protect\\
			E-mail:\{linsihao6@gmail.com\}
\IEEEcompsocthanksitem %Bingqian Lin, 
    Meng Cao is with Peking University, Shenzhen, China. \protect\\
			E-mail:\{mengcaopku@gmail.com\}
\IEEEcompsocthanksitem Jianhua Han and Hang Xu are with Huawei Noah's Ark Lab. \protect\\
 E-mail: hanjianhua4@huawei.com, chromexbjxh@gmail.com.
\IEEEcompsocthanksitem Xiaodan Liang is with Shenzhen Campus of Sun Yat-sen University, Shenzhen, China, Peng Cheng Laboratory, Guangdong Key Laboratory of Big Data Analysis and Processing, Guangzhou, 510006, China.
  \protect\\
  E-mail: liangxd9@mail.sysu.edu.cn
}}
        % <-this % stops a space

\markboth{}%
{TransMamba: Fast Universal Architecture Adaption from Transformers to Mamba}


% \begin{abstract}
% This document describes the most common article elements and how to use the IEEEtran class with \LaTeX \ to produce files that are suitable for submission to the IEEE.  IEEEtran can produce conference, journal, and technical note (correspondence) papers with a suitable choice of class options. 
% \end{abstract}

\IEEEtitleabstractindextext{

\begin{abstract}
Transformer-based architectures have become the backbone of both uni-modal and multi-modal foundation models, largely due to their scalability via attention mechanisms, resulting in a rich ecosystem of publicly available pre-trained models such as LLaVA, CLIP, and DeiT, \etc. In parallel, emerging sub-quadratic architectures like Mamba offer promising efficiency gains by enabling global context modeling with linear complexity. However, training these architectures from scratch remains resource-intensive (\eg, in terms of data and time).
Motivated by this challenge, we explore a cross-architecture knowledge transfer paradigm, termed {\bf \myMethodName}, that facilitates the reuse of Transformer pre-trained knowledge. We propose a two-stage framework to accelerate the training of Mamba-based models, ensuring their effectiveness across both uni-modal and multi-modal tasks. The first stage leverages pre-trained Transformer models to initialize critical components of the Mamba architecture. To bridge architectural and dimensional gaps, we develop a selective weight subcloning strategy and a layered initialization scheme that prioritizes the early $n$ layers.
Building on this initialization, the second stage introduces an adaptive multi-directional knowledge distillation method. This mechanism employs layer-wise adaptive scaling factors to align Mamba representations with their Transformer counterparts, while accommodating the scanning order variations inherent to multi-modal Mamba architectures.
Despite operating with a reduced training dataset and a more compact model architecture, {\myMethodName} consistently outperforms baseline approaches across diverse mamba-based backbones (\eg, PlainMamba, Vmamba, ViM and VideoMamba) and downstream tasks (\eg, image classification, visual question answering, text-video retrieval and multimodal reasoning). All code and implementation details will be released at \href{https://github.com/chen-xw/TransMamba-main}{{https://github.com/chen-xw/TransMamba-main}}.

\end{abstract}

}

\maketitle

\section{Introduction}
\label{sec:intro}

\begin{figure*}[t]
	\centering
        \includegraphics[width=0.99\textwidth]{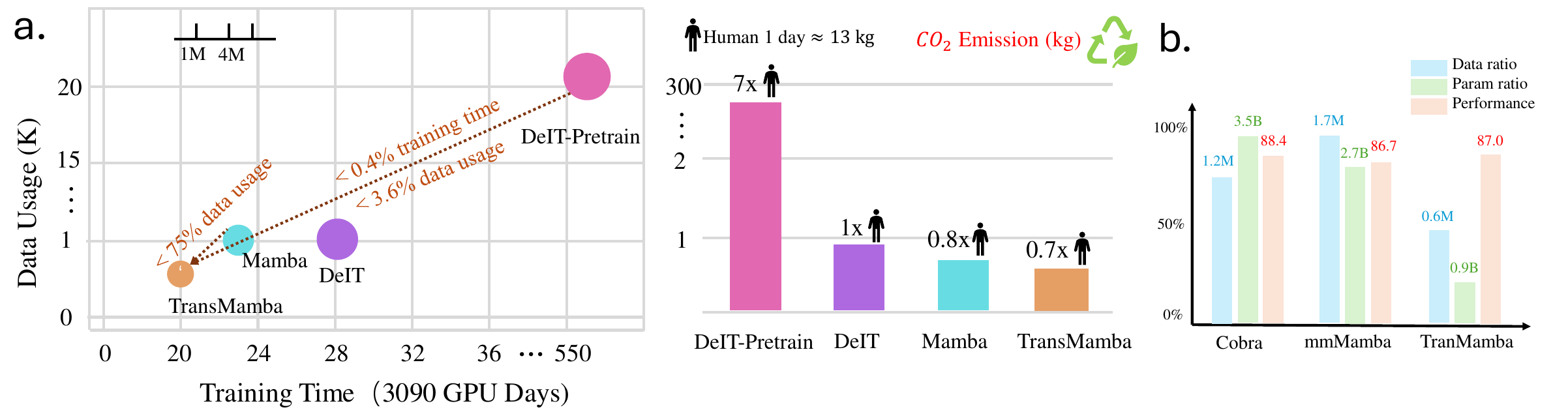}
        %\vspace{-2mm}
        \vspace{-0.2cm}
        \caption{
        \textbf{ (a) Comparisons of training cost (\eg, (left)), $CO_2$ emissions (\eg, (middle)) on uni-modal tasks among different methods. }
        Compared to Mamba, TransMamba uses less data and requires shorter training time.
        \textbf{(b) Comparison of data consumption, parameter count, and model performance across multi-modal task (\eg, (right)).}
        TransMamba achieves performance comparable to existing models while requiring substantially less training data and fewer parameters, as measured by the POPE metric.
        % An overview of \xiuwei{Method Name}. Update in progress, Current Version misunderstanding.
        % \xiuwei{The text is a bit redundant.}}
        }
	\label{fig:pipeline}
	%\vspace{-2mm}
\vspace{-0.4cm}
\end{figure*}

%%%%V1
% Transformer \citep{DBLP:conf/nips/VaswaniSPUJGKP17} architectures have made a transformative impact on the computer vision community \citep{DBLP:conf/cvpr/DongBCZYYCG22,DBLP:conf/iclr/DosovitskiyB0WZ21,DBLP:conf/cvpr/HeZRS16,DBLP:conf/iccv/LiuL00W0LG21,DBLP:journals/corr/SimonyanZ14a},  largely owing to the flexible scalability afforded by attention mechanisms. 
% However, the quadratic complexity of attention mechanism \citep{DBLP:conf/nips/BrownMRSKDNSSAA20} imposes significant computational and memory burdens, which can hinder model optimization and limit scalability in practice.
% % , as demonstrated in Figure \ref{fig:co emossion}.   
% To mitigate these limitations, recent work has introduced sub-quadratic alternatives, such as Mamba \citep{DBLP:journals/corr/abs-2312-00752,DBLP:conf/iclr/FuDSTRR23,DBLP:conf/iclr/GuJTRR23,DBLP:journals/corr/abs-2401-10166,DBLP:journals/corr/abs-2403-17695,DBLP:journals/corr/abs-2401-09417,DBLP:journals/corr/abs-2403-13600,DBLP:journals/corr/abs-2403-06977}, RWKV\citep{peng2023rwkv,duan2024vision} and others. 
% Among these, Mamba is based on the State Space Model (SSM) \citep{DBLP:conf/iclr/GuGR22} and enables global awareness with linear complexity while maintaining competitive performance. 
% % 有点突兀
% Despite their architectural efficiency, training such models (\eg, transformer-like architectures) from scratch remains computationally intensive, particularly when applied to a broad range of downstream tasks, often resulting in considerable energy costs and environmental impact (\cf, Figure \ref{fig:pipeline}).
%%%% V2
Transformer architectures \citep{DBLP:conf/nips/VaswaniSPUJGKP17} have profoundly influenced computer vision \citep{DBLP:conf/iclr/DosovitskiyB0WZ21,DBLP:conf/cvpr/DongBCZYYCG22,DBLP:conf/iccv/LiuL00W0LG21}, largely due to the expressive power and scalability of self-attention. However, their quadratic computational complexity \citep{DBLP:conf/nips/BrownMRSKDNSSAA20} imposes severe memory and efficiency bottlenecks, limiting practical scalability. To address this, recent work has explored sub-quadratic alternatives, most notably Mamba \citep{DBLP:journals/corr/abs-2312-00752,DBLP:conf/iclr/GuJTRR23,11007174}, which is built upon State Space Models (SSMs) \citep{DBLP:conf/iclr/GuGR22} and achieves linear complexity while preserving global context and competitive performance. Despite their efficiency, training such models from scratch across diverse downstream tasks remains resource-intensive, incurring substantial computational costs and environmental impact (\eg, Figure \ref{fig:pipeline}).
% for diverse downstream tasks poses a significant computational burden, as shown in Figure \ref{fig:pipeline} (middle), often resulting in increased energy consumption and carbon footprint.
Fortunately, a wide range of Transformer-based pre-trained models, such as LLaVA \citep{DBLP:conf/icml/RadfordKHRGASAM21},
 CLIP \citep{DBLP:conf/icml/RadfordKHRGASAM21} and
DeiT \citep{DBLP:conf/icml/TouvronCDMSJ21}, have been made publicly available.  
A natural question arises: 
\textit{Can the knowledge embedded in these Transformer-based models be effectively transferred to more efficient sub-quadratic architectures such as Mamba?} 
In this paper, we \textbf{aim to explore this question by developing a cross-architecture transfer paradigm that leverages existing Transformer-based pre-trained models to facilitate the training of  sub-quadratic models}, such as Mamba, in a more computationally efficient and sustainable manner. 

%%%% challenges
Specifically, we address three key challenges in our research: 
1) \textbf{Cross-Architecture Learning}: Effectively adapting pre-trained knowledge from Transformer models to the structurally distinct Mamba architecture, while preserving functional utility and maintaining architectural compatibility, remains a core challenge in cross-framework transfer.
2) \textbf{Multi-Order Dependency in Selective Scanning}: The presence of diverse scanning orders in vision-oriented Mamba models introduces structural divergence, making it difficult to enforce consistency during knowledge distillation across layers.
3) \textbf{Transfer of Multimodal Reasoning Capabilities}:
A relatively underexplored direction that seeks to bridge the gap between the inference strength of multimodal Transformer models and the architectural efficiency of Mamba, particularly as foundation models continue to evolve across modalities.
% 2) \textbf{Equipping SSM-Based Models with Cross-Modal Interaction Capabilities}: This involves developing methods to seamlessly integrate and process information from different modalities, such as text and images, to enhance the versatility and application of SSM-based models in complex tasks.  
% We should ensure that SSM-based models can understand and leverage the relationships between various types of data. 

% \usepackage{multirow}
\begin{table}[ht]
\vspace{-0.1cm}
\centering
\caption{
\textbf{Modality and reasoning support comparison of the proposed method.}
}
\label{tab:aba_difference}
% \vspace{-0.2cm}
\scalebox{0.9}{
\begin{tabular}{ccccc}
\toprule
\multirow{2}{*}{Method} & \multirow{2}{*}{UniModal} & \multicolumn{2}{c}{MultiModal}  & \multirow{2}{*}{Reasoning} \\
                        &                           & Image+Text     & Video+Text   &  \\ \midrule
Hyb-Mamba \citep{DBLP:conf/nips/WangPMRD24}               &   \cmark                        &   \xmark             &    \xmark       & \xmark   \\
Phi-Mamba \citep{DBLP:conf/nips/BickLXKG24}               &   \cmark                        &      \xmark          &     \xmark       & \xmark  \\
mmMamba \citep{DBLP:journals/corr/abs-2502-13145}                 &      \xmark                     &  \cmark              &       \xmark      & \xmark \\
TransMamba(Our)         &    \cmark                       &      \cmark          &    \cmark          & \cmark \\ \bottomrule
\end{tabular}}
% \vspace{-0.2cm}
\vspace{-0.4cm}
\end{table}
% -----------------------------------------------------------------------------------------------
Recent studies have begun to tackle the above challenges from different angles. Representative lines include: 1) \textbf{Direct weight reuse}, 2) \textbf{Progressive distillation}, and 3) \textbf{Hybridization} with a few attention layers kept and the rest replaced by linear-time modules. Most recent methods combine the first two lines and are evaluated primarily on language-only or single vision settings, offering limited evidence of cross-modality generalization, as shown in Table~\ref{tab:aba_difference}. In contrast, our approach couples selective sub-cloning with adaptive multi-directional distillation in a single, modality-agnostic recipe, transferring Transformer knowledge to Mamba across image, image–text, and video–text tasks with reasoning, and thus demonstrating substantially stronger generalization across modalities while preserving Mamba’s linear-time efficiency.

Specifically, we propose a fast and universal two-stage knowledge transfer framework.
In the first stage, Mamba-based architectures (\eg, hybrid and pure variants) are constructed by \textbf{selectively inheriting key weights} from pre-trained Transformer models.
% excluding the QKV projection layers. 
To mitigate architectural discrepancies such as mismatched dimensions and varying layer depths, a selective weight sub-cloning mechanism is introduced alongside a layer-wise initialization strategy, effectively addressing the \textbf{challenge 1)}. Furthermore, convolutional layers intialized with a normal distribution are found to play a stabilizing role during the early phases of training.
The second stage targets \textbf{the alignment of output distributions} between transformer  and mamba architectures. An adaptive multi-directional distillation strategy is employes, assigning layer-specific scaling factors to preserve feature hierarchical feature semantics. To overcome inconsistencies caused by differing scanning orders in vision-oriented mamba variants, additional distillation operations are introduced to explicitly account for scan-specific representations, thereby resolving the issue of \textbf{challenge 2)}. 
%V3
% We propose a fast and universal two-stage knowledge transfer framework to bridge the gap between Transformers and Mamba-based models. In the first stage, we address \textbf{challenge 1)} by constructing Mamba architectures through selective inheritance of pre-trained Transformer weights, augmented with a sub-cloning mechanism and layer-wise initialization to reconcile architectural mismatches. In the second stage, we tackle \textbf{challenge 2)} via an adaptive multi-directional distillation strategy that aligns output distributions while preserving hierarchical semantics and accounting for scan-specific variations in vision Mamba variants.
The framework exhibits strong generalization across a broad range of tasks, \eg, image classification, visual question answering (VQA), and text-video retrieval, as well as diverse Mamba-based model families, \eg, PlainMamba \citep{DBLP:journals/corr/abs-2403-17695}, VMamba \citep{DBLP:journals/corr/abs-2401-10166}, ViM \citep{DBLP:journals/corr/abs-2401-09417} and VideoMamba \citep{DBLP:journals/corr/abs-2403-06977}.
Finally, to address the \textbf{challenge 3)}, an empirical investigation is conducted using multiple publicly available multimodal benchmarks to evaluate the ability of the proposed method to transfer high-level inference skills from transformer models to Mamba architectures.

We claim the following contributions:
\begin{itemize}
    \item  [$ \bullet $] \textbf{Fast and Universal Transfer Framework}: A general two-stage framework is introduced to efficiently transfer knowledge from pre-trained Transformer models to SSM-based architectures, improving training efficiency and downstream performance with minimal overhead.
    \item [$ \bullet $] \textbf{Selective Subcloning Mechanism}: A cross-architecture weight transfer strategy is developed, enabling effective reuse of Transformer parameters by selectively subcloning compatible weights combined with layer-wise initialization, thereby enhancing initialization and convergence.
    \item  [$ \bullet $] \textbf{Adaptive Multi-directional Distillation}: An adaptive multi-directional distillation method is proposed to align features across different scanning orders in vision-oriented Mamba variants, with layer-specific scaling factors to preserve hierarchical feature semantics.
    \item  [$ \bullet $] \textbf{Comprehensive Validation}: The proposed approach is comprehensively validated across diverse mamba-based backbones (\eg, PlainMamba, VMamba, Vim, VideoMamba.) and downstream tasks (\eg, image classification, visual question answering, text-video retrieval, and multimodal reasoning), demonstrating strong generalization and scalability.
    
\end{itemize}

\section{Related Work}

% overview
% Deep neural networks have substantially advanced the research in machine visual perception. % CNN and ViT ? why not use CNN?

% transformer
% \noindent {\bf Vision Transformers} (ViTs) are adapted from the NLP community, showcasing a potent perception model for visual tasks and swiftly evolving into one of the most promissing visual foundation models. Early ViT-based models usually require large-scale dataset and appear in a plain configuration. Later, DeiT~\citep{} employs training techniques to address challenges encountered in the optimization process, and subsequent studies tend to incorporate inductive bias of visual perception into network design. For example, the community propose hierarchival ViTs to gradually decrease the feature resolution throughout the backbone. Moreover, other studies propose to harness the advantages of CNNs, such as introducing convolution operations, designing hybrid architectures by combining CNN and ViT modules, etc.
% disadvantage
% \dx{transformer rather than Vit}

\subsection{Transformers-based models}  
% Since the introduction of 
% Transformers \citep{DBLP:conf/iclr/DosovitskiyB0WZ21,DBLP:conf/cvpr/DongBCZYYCG22,DBLP:conf/iccv/LiuL00W0LG21} 
% % into natural language processing (NLP) \citep{DBLP:journals/aghcs/WrobelKWPW20,DBLP:journals/mta/KhuranaKKS23,DBLP:journals/corr/abs-2304-02017}, they 
% have demonstrated powerful perceptual capabilities for visual tasks and emerged as one of the most promising foundational models for vision. 

\noindent\textbf{- Uni-modal Task.} 
Early ViT-based models typically require large-scale datasets \citep{DBLP:conf/iclr/DosovitskiyB0WZ21} for training and have relatively simple architectures. Later, DeiT~\citep{DBLP:conf/icml/TouvronCDMSJ21} employed training techniques to address challenges encountered in the optimization process, and subsequent research tended to incorporate the inductive biases of visual perception into network design. For instance, the community proposed hierarchical ViTs \citep{DBLP:conf/iccv/LiuL00W0LG21,DBLP:conf/cvpr/DongBCZYYCG22,DBLP:conf/nips/DaiLLT21}, gradually reducing the feature resolution of the backbone network. Additionally, other studies proposed leveraging the advantages of Convolutional Neural Networks (CNNs), such as introducing convolution operations \citep{DBLP:conf/nips/DaiLLT21,DBLP:conf/cvpr/VaswaniRSPHS21} or designing hybrid architectures by combining CNN and ViT modules \citep{DBLP:conf/nips/DaiLLT21}. 
% \xiuwei{

\noindent\textbf{- Multi-modal Task.} CLIP \citep{DBLP:conf/icml/RadfordKHRGASAM21} utilizes multimodal pretraining to redefine classification as a retrieval task, enabling the development of open-domain applications.  
Recently, a number of studies~\citep{liu2024visual,li2024llava,DBLP:journals/corr/abs-2401-02330} have enhanced multimodal capabilities by leveraging powerful large language models (LLMs), typically comprising a vision encoder, a mapper, and an LLM.
LLaVA \citep{liu2024visual} connects CLIP and large language model for end-to-end fine-tuning on generated visual-linguistic instruction data, with excellent performance on multimodal instruction datasets. %%%%
%%%% 添加LLava-next, llava-phi, qwen等有名的模型1
% % Recently, several advanced models have been proposed to further enhance the capabilities of multimodal learning.
% LLaVA-NeXT \citep{li2024llava} enhances LLaVA \citep{liu2024visual} by supporting higher resolutions (e.g., 336×336) and handling multi-image, video and 3D inputs. It uses a more efficient CLIP-ViT-L-336px encoder and a two-layer MLP connection mechanism, 
%and is trained on the M4-Instruct dataset (1.17M samples), demonstrating improved reasoning, OCR, and zero-shot video understanding.
% LLaVA-Phi \citep{DBLP:journals/corr/abs-2401-02330} integrates LLaVA's framework with the 2.7B-parameter Phi-2, a smaller language model, reducing computational complexity while maintaining strong performance across multi-modal benchmarks, notably excelling in ScienceQA \cite{scienceqa}. 
Moreover, Qwen-VL \citep{DBLP:journals/corr/abs-2308-12966} integrates a Vision Transformer encoder, a position-aware adapter, and a large language model, achieving state-of-the-art performance on various benchmarks.
% including image captioning, visual question answering, and expression comprehension.
% PixArt \citep{DBLP:journals/corr/abs-2310-00426}, a transformer-based text-to-image diffusion model, excels in generating high-quality images while reducing training costs and CO$_2$ emissions. 
However, the attention mechanism \citep{DBLP:conf/nips/BrownMRSKDNSSAA20} demonstrates quadratic complexity concerning image token lengths, leading to substantial computational overhead for downstream dense prediction tasks such as object detection \citep{DBLP:journals/pieee/ZouCSGY23}, semantic segmentation \citep{DBLP:journals/eaai/ThisankeDCSVH23}, among others. This limitation curtails the effectiveness of Transformers.

% % Clip  PixArt  
% CLIP \citep{DBLP:conf/icml/RadfordKHRGASAM21} uses multimodel pretraining to convert classification as a retrieval task that enables the pretrained models to transfer knowledge to various downstream tasks. 
% PixArt \citep{DBLP:journals/corr/abs-2310-00426}, a transformer-based text-to-image diffusion model that excels in generating high-quality images while reducing training costs and CO$_2$ emissions.
% However, the attention mechanism \citep{DBLP:conf/nips/BrownMRSKDNSSAA20} requires quadratic complexity in terms of image sizes, resulting in expensive computational overhead when addressing downstream dense prediction tasks such as object detection \citep{DBLP:journals/pieee/ZouCSGY23}, semantic segmentation \citep{DBLP:journals/eaai/ThisankeDCSVH23}, etc. This limits the capability of Transformers.
% }

% Mamba
% S4-->S6--> Vmamba Plainmamba vision mamba(why not use?)
\subsection{SSM-based models} 
State Space Models (SSMs) \citep{DBLP:journals/corr/abs-2312-00752,DBLP:conf/iclr/FuDSTRR23,DBLP:conf/iclr/GuJTRR23,DBLP:journals/corr/abs-2401-10166,DBLP:journals/corr/abs-2403-17695,DBLP:journals/corr/abs-2401-09417,DBLP:journals/corr/abs-2403-13600,DBLP:journals/corr/abs-2403-06977,DBLP:conf/iclr/GuGR22} have proven highly effective in capturing the dynamics and dependencies of language sequences through state space transformations.  
The structured state-space sequence model (S4) \citep{DBLP:conf/iclr/GuGR22,DBLP:conf/nips/GuG0R22} is specifically designed to handle long-range dependencies with linear complexity.  
Following the introduction of S4, more related models have been proposed, such as S5 \citep{DBLP:conf/iclr/SmithWL23}, H3 \citep{DBLP:conf/iclr/FuDSTRR23}, and GSS \citep{DBLP:conf/iclr/Mehta0CN23}.  
Mamba stands out by incorporating a data-dependent SSM layer and a selection mechanism known as the parallel scan (S6) \citep{DBLP:journals/corr/abs-2312-00752}. 
Compared to Transformer-based models, which rely on attention mechanism with quadratic-complexity, Mamba excels at processing long sequences with linear complexity.  
% In computer vision, SSM was first applied to pixel-level image classification, while S4 was utilized to manage long-range temporal dependencies in movie clip classification. 
% In compute vision,
% Moreover, Mamba's potential has spurred numerous studies, showcasing its superior performance and higher GPU efficiency over transformers in visual tasks like object detection \citep{DBLP:journals/pieee/ZouCSGY23} and semantic segmentation \citep{DBLP:journals/corr/abs-2401-04722}.  
% Distinct from previous works, our TransMamba aims to explore the potential of using knowledge from pre-trained Transformer models to a new model with Mamba architecture in a cross-architecture transferring way.
%%%% 添加Vision mabma, Vmamba, PlainMamba，Cobra, videoMamba等常见mamba结构论文
% (SSMs) are recently proposed models that are introduced into deep learning as state space transforming. 
Recently, ViM \citep{DBLP:journals/corr/abs-2403-06977}, VMamba \citep{DBLP:journals/corr/abs-2401-10166} and PlainMamba \citep{DBLP:journals/corr/abs-2403-17695} has introduced the Mamba architecture into the vision domain by designing a multi-directional selection mechanism to effectively handle the two-dimensional structure of images.
%  uses bidirectional state space block with positional embeddings to capture the global visual context. 
% VMamba \citep{DBLP:journals/corr/abs-2401-10166} extends one-dimensional scanning to a four-directional 2D selective scan, efficient spatial modeling. 
% PlainMamba \citep{DBLP:journals/corr/abs-2403-17695} utilizes continuous 2-D scanning with direction-aware updates to preserve consistent feature resolution for visual recognition.
% Further, VideoMamba \citep{DBLP:journals/corr/abs-2403-06977} extends Mamba with 3D causal convolutional SSMs, achieving linear complexity with respect to frame count and outperforming Transformer-based counterparts like ViViT \citep{DBLP:journals/corr/abs-2103-15691} .
Moveover, in multi-modal scenarios, Cobra \citep{DBLP:journals/corr/abs-2403-14520} replaces the attention layers with SSM blocks in large vision-language models, preserving cross-modal reasoning ability while retaining linear sequence complexity. 
Hybrid architectures such as Jmamba-1.5 \citep{team2408jamba} and MambaOut \citep{DBLP:journals/corr/abs-2405-07992} combine Mamba with lightweight attention or convolutional modules to balance efficiency and performance across varying context lengths and high-resolution inputs.
Distinct from previous works, our TransMamba aims to explore the potential of using knowledge from pre-trained Transformer models to a new model with Mamba architecture in a cross-architecture transferring way.
% Although these architectures achieve efficient inference, their large-scale pre-training remains costly. Recent NLP studies have shown that reusing Transformer weights can accelerate convergence, yet systematic weight transfer across heterogeneous depths, dimensions, and multi-directional scanning orders has not been investigated for vision or multi-modal tasks. 
% To fill this gap, we propose TransMamba, which adapts pre-trained Transformer knowledge to Mamba via selective weight sub-cloning and multi-directional distillation, preserving linear complexity while aligning direction-specific features for image, image-text, and video-text applications.
% distillation
\subsection{Transfer learning}
\noindent\textbf{- Traditional distillation methods.}
% 几篇代表性工作即可，大约页面的1/3填满
Knowledge Distillation (KD)~\citep{hinton2015distilling,10198386,yang2024vitkd,pham2024frequency,saadi2025class} is an effective technique for model compression and knowledge transfer. 
\citep{hinton2015distilling} first introduced the concept of using soft targets from a teacher model as supervision signals to guide the training of a student model, a method known as response-based distillation.
To convey richer intermediate information to the student model, researchers proposed feature-based distillation\citep{yang2024vitkd, pham2024frequency, saadi2025class,liu2024rethinking}. 
These methods enhance student learning by matching feature maps from intermediate layers of the teacher and student models, providing stronger guidance, and achieve strong performance across multiple visual tasks.
% Building on feature-based distillation methods, relation-based distillation was introduced\citep{feng2024relational,xiao2024boosting,park2019relational}. Instead of focusing on the features themselves, it emphasizes the relationships between features, such as attention maps (Attention Transfer) or inter-instance relation graphs, thereby enabling the student to learn higher-level structured knowledge and significantly expanding the applicability and effectiveness of knowledge distillation.
However, these traditional knowledge distillation methods focus on knowledge transfer between homogeneous architectures, overlooking the potential challenges posed by distillation across heterogeneous architectures.

\noindent\textbf{- Transformer to Mamba distillation.}
Recently, some work \citep{wang2024mamba,bick2024transformers} in the NLP field has focused on the knowledge transfer process from Transformers to Mamba.
%%%%V1
% \citep{wang2024mamba} proposes to reuse the linear projection weights of the attention layer of the large transformer model to perform cross-architecture distillation with less GPU resources, achieving performance comparable to that of the large transformer model.
% \citep{bick2024transformers} views both Transformers and SSMs as applying different forms of mixing matrices over the token sequences, and propose a progressive distillation strategy to distill the Transformer architecture by matching different degrees of granularity in the SSM.
% \citep{lei2024dvmsr} use a simple $l_1$ distillation loss for leveraging the rich representation knowledge of teacher network.
%%%%V2
~\citep{wang2024mamba} reuses Transformer attention weights for cross-architecture distillation with reduced GPU cost. ~\citep{bick2024transformers} unifies Transformers and SSMs via token mixing and proposes progressive distillation across granularities. \citep{lei2024dvmsr} uses an $l_1$ loss to transfer rich representations.
% 添加Multimodal Mamba: Decoder-only Multimodal State Space Model via Quadratic to Linear Distillation与其他论文
% \citep{Liao2025MultimodalMD} proposes a hybrid architecture combining Transformer and Mamba, enabling efficient cross-architecture knowledge transfer and retention of multimodal capabilities by inheriting pre-trained Transformer structural parameters and employing a three-stage distillation strategy.
\citep{Liao2025MultimodalMD} proposes a Transformer-Mamba hybrid for efficient knowledge transfer and multimodal capability retention via parameter inheritance and three-stage distillation.
% \citep{} 
While knowledge transfer between architectures has been extensively studied, limited attention has been given to transferring knowledge from Transformers to Mamba, particularly in the context of vision and multimodal tasks. The incorporation of visual information further complicates the Mamba architecture, presenting additional challenges for effective transfer. In this work, we address these challenges by investigating efficient knowledge transfer strategies from Transformers to Mamba across both uni-modal and multi-modal settings.
\section{Method}
\label{sec:met}

%%%%%%%%Overall structure, Transmamba, details for different tasks

%-----------------------------------------------------------------------
\begin{figure*}[t]
  \centering
   \includegraphics[width=0.98\linewidth]{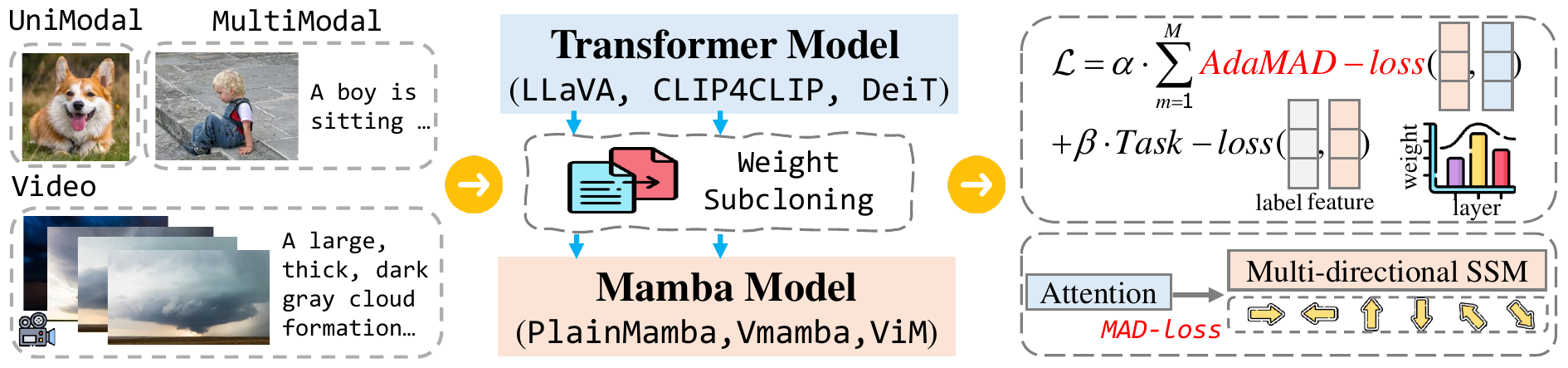}
   \caption{{\bf Overview of the main components of our \myMethodName.} It supports uni-modal, multi-modal, and video inputs through a two-stage cross-architecture transfer process. The {\bf middle} stage illustrates selective weight subcloning, enabling partial inheritance of Transformer-based parameters. The {\bf right} stage depicts the adaptive multi-directional feature distillation, which combines layer-wise weighting with direction-aware distillation tailored to the multi-directional scanning structures in visual Mamba architectures. }
   \label{fig:main-architecture}
\vspace{-0.6cm}
\end{figure*}
%-----------------------------------------------------------------------

% This work explores the feasibility of cross-architecture knowledge transfer between quadratic and sub-quadratic sequence models. Taking Mamba as a representative sub-quadratic architecture, we investigate how to effectively transfer knowledge from pre-trained Transformer models to Mamba in a cost-efficient and robust manner. We begin by outlining the core principles of state space models (SSMs), then introduce our training framework, \myMethodName. We further detail the adaptation strategies employed within \myMethodName for both uni-modal and multi-modal tasks, along with a comprehensive description of the overall training pipeline.

% background
\subsection{Preliminary} %\xw{Not goog enough! rewrite!}} 
\label{subsec:pre}
State Space Models (SSMs) are constructed upon continuous systems that translate a 1D function or sequence, \(x(t) \in \mathbb{R}^L \rightarrow y(t) \in \mathbb{R}^L\), using a hidden state \(h(t) \in \mathbb{R}^N\). Formally, SSMs employ the following ordinary differential equation (ODE) to describe the input data:

\begin{equation}
\begin{aligned}
h'(t) &= \mathbf{A}h(t) + \mathbf{B}x(t), \\
y(t) &= \mathbf{C}h(t),
\end{aligned}
\label{eq:ssm}
\end{equation}
where \(\mathbf{A} \in \mathbb{R}^{N\times N}\) signifies the system's evolution matrix, and \(\mathbf{B} \in \mathbb{R}^{N\times 1}\), \(\mathbf{C} \in \mathbb{R}^{N\times 1}\) represent the projection matrices. This continuous ODE is approximated through discretization in modern SSMs. Mamba is one of the discrete versions of the continuous system, integrating a timescale parameter \(\mathbf{\Delta}\) to transform the continuous parameters \(\mathbf{A}, \mathbf{B}\) into their discrete counterparts \(\overline{\mathbf{A}}, \overline{\mathbf{B}}\).  
The typical method for this transformation involves employing the zero-order hold (ZOH) method, defined as:

\begin{equation}
\begin{aligned}
\overline{\mathbf{A}} &= \exp(\mathbf{\Delta \mathbf{A}}), \\
\overline{\mathbf{B}} &= (\mathbf{\Delta \mathbf{A}})^{-1} (\exp(\mathbf{\Delta \mathbf{A}}) - \mathbf{I}) \cdot \mathbf{\Delta \mathbf{B}}, \\
h_t &= \overline{\mathbf{A}} h_{t-1} + \overline{\mathbf{B}} x_t, \\
y_t &= \mathbf{C}h_t,
\end{aligned}
\label{eq:discrete ssm}
\end{equation}
where the parameters \(\mathbf{B} \in \mathbb{R}^{B\times L \times N}\), \(\mathbf{C} \in \mathbb{R}^{B\times L \times N}\), and \(\mathbf{\Delta} \in \mathbb{R}^{B \times L \times D}\). 
Contrary to conventional models relying on linear time-invariant SSMs, Mamba distinguishes itself by integrating a Selective Scan Mechanism (S6) as its core SSM operator. More precisely, three functions ${S_C}(x)$, ${S_B}(x)$, ${S_\Delta}(x)$ are introduced to associate parameters $\bar B$, $C$, $\Delta$ in Equation \ref{eq:discrete ssm} to the input data $x$. Based on ${S_\Delta}(x)$, $\bar A$ can also be associated with the input data $x$.
When given an input sequence $X: = [{x_1}, \cdots ,{x_N}] \in {R^{N \times D}}$ of $N$ feature vectors, The ouput sequence $Y$ can be denoted as:
% \begin{equation}
% \begin{aligned}
% { \tiny
% Y = \left[ {\begin{array}{*{20}{c}}
% {{{\bar C}_1}}&0& \cdots &0\\
% 0&{{{\bar C}_2}}& \cdots &0\\
%  \vdots & \vdots & \ddots & \vdots \\
% 0&0& \cdots &{{{\bar C}_N}}
% \end{array}} \right]\left[ {\begin{array}{*{20}{c}}
% {{{\bar B}_1}}&0& \cdots &0\\
% {{{\bar A}_2}{{\bar B}_1}}&{{{\bar B}_2}}& \cdots &0\\
%  \vdots & \vdots & \ddots & \vdots \\
% {{{\bar A}_N} \cdots {{\bar A}_2}{{\bar B}_1}}&{{{\bar A}_N} \cdots {{\bar A}_3}{{\bar B}_2}}& \cdots &{{{\bar B}_N}}
% \end{array}} \right]\left[ {\begin{array}{*{20}{c}}
% {{x_1}}\\
% {{x_2}}\\
%  \vdots \\
% {{x_N}}
% \end{array}} \right] }
% \end{aligned}
% \label{eq:output}
% \end{equation}
\begin{equation}
\begin{aligned}
{ \tiny
Y = C\left[ {\begin{array}{*{20}{c}}
{{{\bar B}_1}}&0& \cdots &0\\
{{{\bar A}_2}{{\bar B}_1}}&{{{\bar B}_2}}& \cdots &0\\
 \vdots & \vdots & \ddots & \vdots \\
{{{\bar A}_N} \cdots {{\bar A}_2}{{\bar B}_1}}&{{{\bar A}_N} \cdots {{\bar A}_3}{{\bar B}_2}}& \cdots &{{{\bar B}_N}}
\end{array}} \right]\left[ {\begin{array}{*{20}{c}}
{{x_1}}\\
{{x_2}}\\
 \vdots \\
{{x_N}}
\end{array}} \right] }
\end{aligned}
\label{eq:output}
\end{equation}
which can be expressed as: $Y=C(MX)$. 

\subsection{TransMamba}

\noindent {\bf - Selective Subcloning Mechanism.}
% ----------------------------------------------------------------------------
\begin{figure}[t]
  \centering
   \includegraphics[width=0.98\linewidth]{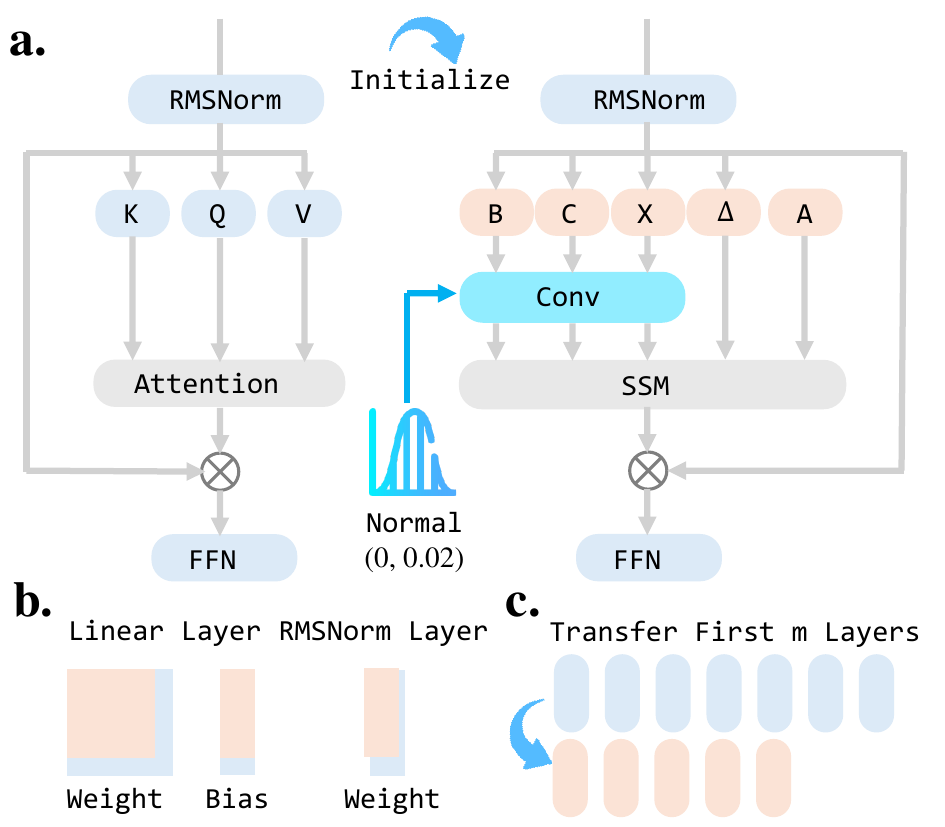}

   \caption{{\bf Overview of the Selective Subcloning Mechanism.}
   % (a) For multimodal tasks, partial Transformer blocks are inherited, while convolutional components are initialized with a regularized distribution. (b) For structural mismatches in dimensionality, partial weight subcloning is applied by mapping overlapping submatrices. (c) For differences in model depth, the first $n$ layers of the Transformer are used to initialize the target Mamba model. 
   }
   % \xw{Remember to show Repeated Initialization of Transformer Model Structure.}}
   \label{fig:fig-weight subcloning}
\vspace{-0.4cm}
\end{figure}
% ----------------------------------------------------------------------------
To effectively harness pre-trained transformer within a structurally distinct mamba architecture, we introduce a targeted initiliazation stragegy. Specifically, we replace the attention module with the mamba block while preserving the original transformer's feed-forward (\eg, FFN) and normalization (\eg, RMSNorm) layers. 
Parameters for mamba block are initialized using original mamba weights (\cf, Figure \ref{fig:fig-weight subcloning} (a)).
For differences in layer depth, we initialize the mamba layers using the parameters from the first $n$ layers of the transformer model (where $n$ is the number of Mamba layers), as deeper Transformer layers tend to encode task-specific or high-level semantic information that may hinder the early training of Mamba (\cf, Figure \ref{fig:fig-weight subcloning} (c)). 
For structural mismatches in parameter dimensions, inspired by~\cite{Samragh2023WeightSD}, we introduce a partial weight initialization method (\cf, Figure \ref{fig:fig-weight subcloning} (b)), where overlapping submatrices from pre-trained weights are mapped to the corresponding regions of the target architecture. For example, linear weights are initialized along their top-left diagonal sub-block, and bias or RMSNorm parameters are partially copied based on dimensional compatibility.
This initialization scheme enables maximal reuse of transferable knowledge while preserving compatibility. Empirically, we observe that such carefully conditioned initialization, particularly for convolutional components, is critical for training stability. Naive initialization often leads to collapse, whereas regularized initialization significantly improves convergence behavior.

\noindent {\bf - Adaptive Multi-Direction Distillation.}
% -----------------------------------------------------------------------------------------------
\begin{figure}[t]
  \centering
   \includegraphics[width=0.8\linewidth]{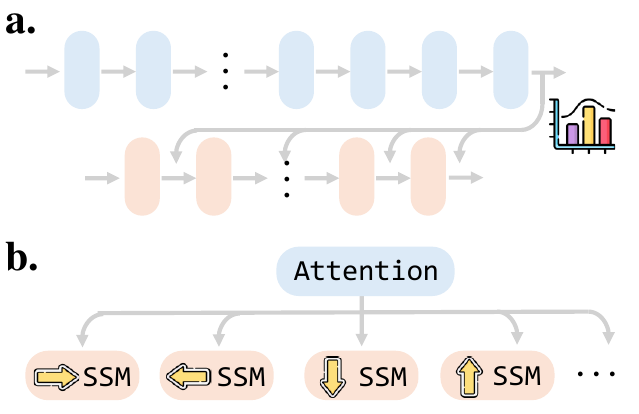}

   \caption{{\bf Overview of the Adaptive Multi-Direction Distillation.}  
   % (a) Adaptive layer distillation, where the final-layer features of the Transformer are used to supervise all layers of the Mamba model with layer-specific weights. (b) Multi-directional distillation tailored to the diverse scanning orders inherent in different visual Mamba variants.
   }
   \label{fig:layer-distill}
\vspace{-0.6cm}
\end{figure}
% -----------------------------------------------------------------------------------------------
To facilitate effective cross-architecture knowledge transfer, we adopt a knowledge distillation framework that transfers representational patterns (\eg, $F$) from a transformer-based teacher model $\mathcal{T}$ to a mamba-based student model $\mathcal{S}$. 
% 层数
Our objective is to bridge architectural differences while preserving the rich feature distributions embedded in the transformer.
We consider two scenarios based on the alignment of layer depth between the two models. When the number of layers is matched, we align the intermediate features layer-wise using cosine similarity loss (\cf, Equation \ref{eq:loss_distill_similar}).
\begin{equation}
{\mathcal{L}_{distill\_1}} = \sum\limits_{i = 1}^N {(1 - \cos (\theta_i ))} = \sum\limits_{i = 1}^N{(1 - \frac{{F_{\mathcal{T}\_i}F_{\mathcal{S}\_i}}}{{||F_{\mathcal{T}\_i}||\cdot||F_{\mathcal{S}\_i}||}})},
\label{eq:loss_distill_similar}
\end{equation}
where $\theta_i$ is the angle between the $i$-th layer outputs of the teacher and student models, ${F_{\mathcal{T}\_i}}$ denotes the feature output of the $i$-th layer of the teacher model $\mathcal{T}$.
In the case of mismatched depths, we regularize all student layers using the final-layer representation of the teacher (\cf, Equation \ref{eq:loss_distill_different}). 
\begin{equation}
{\mathcal{L}_{distill\_2}} = \sum\limits_{i = 1}^N {(1 - \cos (\theta_i ))} = \sum\limits_{i = 1}^N{(1 - \frac{{F_\mathcal{T}F_{\mathcal{S}\_i}}}{{||F_\mathcal{T}||\cdot||F_{\mathcal{S}\_i}||}})},
\label{eq:loss_distill_different}
\end{equation}
${F_{\mathcal{T}}}$ denotes the feature output of the last layer of the teacher model $\mathcal{T}$.
% To mitigate dimensional inconsistencies, we introduce a lightweight linear projection layer that aligns the teacher's feature space to that of the student prior to loss computation.

However, naively enforcing uniform distributional constraints across layers may induce optimization imbalance and hinder convergence. To address this, we introduce an adaptive distillation weighting mechanism (\cf, Equation \ref{eq:loss_adapt_distill}), where a parameter-sharing meta-network dynamically predicts layer-wise importance weights, enabling self-regulated and balanced learning during distillation.
\begin{equation}
{\mathcal{L}_{\text{AdaptDistill}}} = \sum\limits_{i = 1}^N {\text{Adapt-factor}\cdot(1 - \cos (\theta_i ))}
\label{eq:loss_adapt_distill}
\end{equation}

Furthermore, we highlight an important architectural distinction: unlike transformers that operate over global self-attention matrices, the structured state space design of vision-oriented mamba introduces direction-specific dependencies. The scanning order of image patches significantly affects 2D spatial modeling. While prior work(\eg, ViM, Vmamba, etc.) explores multiple scanning orders, we introduce primarily on the bidirectional representation as a key structural paradigm. Its forward and backward computations produce asymmetrical, simplified the output form of bidirectional Mamba as follows: 
\begin{equation}
{Y_{\text{{forward}}}} = C\left[ {\begin{array}{*{20}{c}}
{{f_{11}}}&0&0\\
{{f_{21}}}&{{f_{22}}}&0\\
{{f_{31}}}&{{f_{32}}}&{{f_{33}}}
\end{array}} \right]\left[ {\begin{array}{*{20}{c}}
{{x_1}}\\
{{x_2}}\\
{{x_3}}
\end{array}} \right]
\label{eq:forward}
\end{equation}
\begin{equation}
{Y_{\text{{backward}}}} = C\left[ {\begin{array}{*{20}{c}}
{{b_{33}}}&0&0\\
{{b_{23}}}&{{b_{22}}}&0\\
{{b_{13}}}&{{b_{12}}}&{{b_{11}}}
\end{array}} \right]\left[ {\begin{array}{*{20}{c}}
{{x_3}}\\
{{x_2}}\\
{{x_1}}
\end{array}} \right]
\label{eq:backward}
\end{equation}
Then the lower-triangular matrices merged into dense representations.
\begin{equation}
Y = C\left[ {\begin{array}{*{20}{c}}
{{f_{11}} + {b_{33}}}&{{b_{12}}}&{{b_{13}}}\\
{{f_{21}}}&{{f_{22}} + {b_{22}}}&{{b_{23}}}\\
{{f_{31}}}&{{f_{32}}}&{{f_{33}} + {b_{33}}}
\end{array}} \right]\left[ {\begin{array}{*{20}{c}}
{{x_1}}\\
{{x_2}}\\
{{x_3}}
\end{array}} \right]
\label{eq:last}
\end{equation}
Compared to the standard Transformer formulation $Y = SV = (SX) V$, where $S = soft\max (Q{K^{\rm T}}/\sqrt D )$, the bi-directional mamba representation $Y = C (MX)$ exhibits repeated diagonal elements in the transition matrix $M$, as also observed in VideoMamba \citep{DBLP:journals/corr/abs-2403-06977}.
This structural divergence may lead to inconsistent feature alignment if treated uniformly.
To address this, we propose a {\bf multi-directional knowledge distillation method} that explicitly decouples the direction-specific features during distillation.
This design mitigates the over-optimization or under-optimization of redundant components, ensuring more faithful and stable knowledge transfer between heterogeneous architectures.
%%%% 双向蒸馏过程
%% 前向过程
For the forward process, we directly leverage the aligned transformer output features as the supervision signals (\cf, Equation \ref{eq:forward loss}).
\begin{equation}
{\mathcal{L}_{\text{{forward}}}} = {\text{AdaptDistill}}(F_\mathcal{T} ,F_{\mathcal{S}\_forward} )
\label{eq:forward loss}
\end{equation}
%%%% 后向过程，需不需要加mask呢？去除重复元素
For the backward process, we apply a reverse-aware projection layer on the transformer outputs to align them  with the backward-structure of the mamba transition matrix (\cf, Equation \ref{eq:backward loss}).
\begin{equation}
{\small
{\mathcal{L}_{\text{backward}}} = {\text{AdaptDistill}}({\text{Reverse}}(F_\mathcal{T}) ,F_{\mathcal{S}\_backward} )}
\label{eq:backward loss}
\end{equation}
The distillation strategy for other scanning orders follows a similar methodology. 
Consequently, for any given task, the overall loss function is defined as Equation \ref{eq:loss_all}.
\begin{equation}
 \mathcal{L}_{\text{total}} =  \alpha \mathcal{L}_{\text{task}}+ (1-\alpha) (\mathcal{L}_{\text{forward}} + \mathcal{L}_{\text{backward}}), 
\label{eq:loss_all}
\end{equation}

\subsection{Downstream Tasks.} 
To validate the effectiveness of \textbf{TransMamba}, we consider both uni-modal and multi-modal tasks, including image classification, visual question answering, and text-video retrieval.

\noindent\textbf{- Uni-modal Task.} 
For image classification, we employ three state-of-the-art Mamba architectures as the student models, \eg, Vmamba \cite{DBLP:journals/corr/abs-2401-10166}, PlainMamba \cite{DBLP:journals/corr/abs-2403-17695}, and ViM \cite{DBLP:journals/corr/abs-2401-09417}. Knowledge transfer is performed from pre-trained Transformer DeiT models (ImageNet1k/21k) \cite{DBLP:conf/icml/TouvronCDMSJ21} to these Mamba-based students. Due to substantial architectural differences, the selective weight sub-cloning stage is generally omitted for main single-modality tasks. Our main experiments leverage the adaptive multi-directional distillation framework. Notably, the selective weight subcloning is applied to PlainMamba, where it accelerates early-stage convergence with minimal impact on final performance (\cf, Figure~\ref{fig:accuracy} (d)).

\noindent\textbf{- Multi-modal Task.} 
Inspired by the LLaMA architecture, we design two novel mamba-based architectures: Llama-Mamba and Llama-Mamba-Hybrid. Both replace attention modules in LLaMa with state space model (SSM) blocks, while retaining feed-forward network (\eg, FFN) and normalization (\eg, RMSNorm) layers unchanged. Llama-Mamba substitutes all attention layers with SSM-based Mamba layers, whereas lama-Mamba-Hybrid adopts a partial replacement strategy, substituting varying proportions (\eg, 1/8, 1/4, 1/2, interleaved, \etc) of attention layers at different network depths (\eg, early, middle, and final).
Training large mamba-based models exhibits gradient instability consistent with prior NLP studies in \citep{zuo2024falcon,team2408jamba}. 
To ensure stability, we inherit FFN and RMSNorm weights directly from LLaMa, while initializing architecture-specific linear layers with weights from a standard Mamba model~\footnote{https://huggingface.co/state-spaces/models} .
Our experiments reveal that the initialization of convolutional layers significantly affects training stability; thus, we apply reparameterized initialization from a normalized distribution to improve training robustness.
Using above architectures, we employ an adaptive distillation framework to align layer-wise feature distributions between Transformer teachers and Mamba students (\cf, Figure~\ref{fig:layer-distill}), facilitating effective cross-architecture knowledge transfer while preserving representational consistency.
For text-video retrieval, we inherit SSM parameters from VideoMamba \cite{DBLP:journals/corr/abs-2403-06977}, providing a straightforward but effective demonstration of our method’s generalizability to video-based tasks.

% 最后到底要用哪个数据集？llava-cot没有PRM啊，xiangkun的数据集太多了，有没有其他可替代数据？
\subsection{Multimodal Reasoning.} 
To investigate whether reasoning capabilities from transformer models can be transferred to mamba-based architectures, we first collect approximately 80K samples from LLaVA-CoT \citep{xu2024llava} and fine-tune a reasoning-capable transformer model based on llama architecture using this dataset. 
The construction of a structured reasoning format is pivotal for enhancing downstream reasoning capabilities. We adopt a framework comprising \texttt{<SUMMARY></SUMMARY>}, \texttt{<CAPTION></CAPTION>}, \texttt{<REASONING></REASONING>}, and \texttt{<CONCLUSION></CONCLUSION>} tags, following the design principles of LLaVA-CoT \citep{xu2024llava}. This structured representation fosters coherent, interpretable, and logically consistent reasoning throughout the inference process.
Subsequently, we curate an additional 40K dataset and employ the distillation method described above to transfer knowledge from the llama model into a mamba architecture. 
The distillation strategy follows a scheme analogous to that used in multimodal tasks, ensuring effective transfer of reasoning behavior while accommodating architectural differences.

\section{Experiments}
\label{sec:exp}
% We conduct experiments to show the effectiveness of our \myMethodName. Firstly, we introduce the datasets and evaluation metrics in Section \ref{sec:4.1}. Then, in Section \ref{sec:4.3}, we compare the proposed \myMethodName with other advanced methods. In Section \ref{sec:4.4} we perform extensive ablation studies to evaluate the contribution of each design in our \myMethodName. Finally, we visualize and discuss the output examples of our \myMethodName for better understanding in Section \ref{sec:4.5}.

% We focus on four scenes in the experiments, including image classification, visual question answering, text-video retrieval and multimodal reasoning, as mentioned in Section \ref{sec:met}.

\subsection{Datasets and Evaluation Metrics}\label{sec:4.1}
% In this section, we introduce the datasets and evaluation metrics for the corresponding models.
\noindent {\bf - Datasets of single model. } 
% \textbf{Single model.}
For image classification, we evaluate our approach on three benchmark datasets: CIFAR-100 \citep{krizhevsky2009learning}, ImageNet-100 \citep{deng2009imagenet} (\eg, ImageNet$^S$), and ImageNet-1000 \citep{deng2009imagenet}. CIFAR-100 contains 100 classes with 600 images each, split into 500 training and 100 testing samples per class. ImageNet1000 \citep{deng2009imagenet} is a large-scale dataset
 with 1,000 classes that includes $1.28$ million images
 for training and 50,000 images for validation. ImageNet-100 is a subset of 100 randomly selected classes from the full ImageNet-1000 dataset.

\noindent \textbf{- Datasets of multi model (\eg, visual question answering (VQA)).}
For visual question answering, we utilize 
% $558$k general captioning samples from the LLaVA-1.5-pretrain datasets and 
$665$k captioning and conversation samples from LLaVA-1.5-finetune \citep{liu2024visual} datasets. 
LLaVA-1.5-finetune datasets include LLaVA \citep{liu2024visual}, VQAv2 \citep{DBLP:conf/cvpr/GoyalKSBP17}, GQA \citep{hudson2019gqa}, OKVQA \citep{DBLP:conf/cvpr/MarinoRFM19}, OCRVQA \citep{DBLP:conf/icdar/0001SSC19}, AOKVQA \citep{DBLP:conf/eccv/SchwenkKCMM22}, TextCaps \citep{DBLP:conf/eccv/SidorovHRS20}, RefCOCO \citep{DBLP:conf/emnlp/KazemzadehOMB14} and VG \citep{DBLP:journals/ijcv/KrishnaZGJHKCKL17}.
% LLaVA \citep{} 
% ShareGPT \citep{}
% VQAv2 \citep{}
% GQA \citep{}
% OKVQA \citep{}
% OCRVQA \citep{}
% AOKVQA \citep{}
% TextCaps \citep{}
% RefCOCO \citep{}
% VG \citep{}

\noindent \textbf{- Datasets of multi model (\eg, text-video retrieval).}
For text-video retrieval, we employ two benchmark datasets: MSR-VTT \citep{DBLP:conf/cvpr/XuMYR16} and DiDeMo \citep{DBLP:conf/iccv/HendricksWSSDR17}. MSR-VTT is a large-scale open-domain video captioning dataset comprising 10,000 clips spanning 20 categories, each annotated with 20 English sentences via Amazon Mechanical Turk. DiDeMo consists of 8,395 training, 1,065 validation, and 1,004 test videos.

\noindent \textbf{- Datasets of multimodal reasoning.}
We utilized the multimodal reasoning data from the LLaVA-CoT~\citep{xu2024llava} paper,
which comprises components ChartQA \cite{chartqa}, A-OKVQA \cite{a-okvqa}, DocVQA \cite{docvqa}, PISC \cite{pisc}, CLEVR \cite{clevr}, GeoQA+ \cite{geoqa+}, AI2D \cite{ai2d}, ScienceQA \cite{scienceqa}, CLEVR-Math \cite{clevr-math}.
% ShareGPT4V \cite{sharegpt4v} provides multi-turn visual question-answering data from GPT-4V interactions.
% ChartQA \cite{chartqa} focuses on the interpretation of charts and graphical information.
% A-OKVQA \cite{a-okvqa} emphasizes reasoning that requires external knowledge beyond what is visually present.
% DocVQA \cite{docvqa} involves answering questions based on document content and textual understanding.
% PISC \cite{pisc} aims to evaluate the understanding of social relationships in images.
% CLEVR \cite{clevr} targets reasoning about object properties, spatial relations, and counting.
% GeoQA+ \cite{geoqa+} is designed for geometric reasoning tasks.
% AI2D \cite{ai2d} focuses on scientific diagrams and associated comprehension tasks.
% ScienceQA \cite{scienceqa} addresses question-answering in scientific contexts.
% CLEVR-Math \cite{clevr-math} extends CLEVR with arithmetic reasoning in visual settings.

\noindent \textbf{- Evaluation metrics.} 
For {\textbf{classification task}, we employ the top accuracy as the evaluation metric. 
For {\textbf{visual question answering}, we conduct experiments on 
POPE \citep{DBLP:conf/emnlp/LiDZWZW23}, MME \citep{Fu2023MMEAC}, MMB \citep{liu2024mmbench}, Seed \citep{li2023seed} and MMMU \citep{yue2024mmmu}.
% GQA\citep{hudson2019gqa}, VizWiz\citep{gurari2018vizwiz}, TextVQA\citep{singh2019towards}, MME\citep{Fu2023MMEAC}.
% VQA-v2 \citep{DBLP:conf/cvpr/GoyalKSBP17} and GQA \citep{hudson2019gqa} evaluate model’s visual perception capabilities on open-ended short answers. 
% VizWiz \citep{gurari2018vizwiz} contains 8,000 images to evaluate model’s zero-shot generalization on visual questions asked by visually impaired people.
POPE \citep{DBLP:conf/emnlp/LiDZWZW23} evaluates model’s degree of hallucination on three sampled subsets of COCO \citep{DBLP:conf/eccv/LinMBHPRDZ14}: random, common, and adversarial and we report the F1 score on all three splits.
% MME-Perception \citep{Fu2023MMEAC} focuses on fine-grained visual perception and reasoning, probing models via binary true/false questions grounded in image content.
MMB \citep{liu2024mmbench} assesses robustness and consistency across diverse visual contexts through carefully curated multiple-choice questions.
For open-world, compositional understanding, Seed \citep{li2023seed} evaluates models’ ability to reason over unconstrained, real-world multimodal inputs.
At the higher end of cognitive demand, MMMU \citep{yue2024mmmu} challenges models with college-level, cross-disciplinary reasoning tasks requiring deep integration of visual and textual knowledge.
For {\textbf{text-video retrieval task}, we assess model performance using recall@k and Mean R.
For \textbf{Multimodal Reasoning}, MathVista\citep{Lu2023MathVistaEM} and MMMU\citep{yue2024mmmu} are conducted in our experiments. 
For MathVista, we use the Test Mini split (\eg, around 1,000 samples). For MMMU, we use the val dataset. \textbf{Note:} This evaluation is based on strict matching without GPT-based answer filtering, and thus the results are expected to be lower compared to those obtained with GPT-filtered evaluation.

%-----------------------------------------------------------------------------------------------
\begin{table}[]
\centering
\caption{
{\bf Architectural comparison of depth, channels, and parameter scale on image classification.}
}
\label{tab:variants}
\begin{tabular}{cccc}
\toprule
% \midrule
Model    & Depth & Channels & Params \\ \midrule
VMamba-T & 15    &  /      & 40.5M   \\
PMamba-T & 12    & 192      & 4.1M   \\
PMamba-S & 12    & 384      & 13.7M  \\
PMamba-B & 12    & 768      & 49.5M  \\ 
ViM-T & 12    & 192      & 7.0M   \\
ViM-S & 12    & 384      & 25.5M  \\

\midrule
\end{tabular}
\vspace{-0.4cm}
\end{table}

\begin{table*}[]
\centering
\caption{{\bf
Comparison with state-of-the-art MLLMs on the commonly-used multimodal benchmarks for MLLMs.}
% “Recipe” denotes the adopted training recipe. 
% “PT”, “SFT”, and “DT” denote the pre-training, supervised fine-tuning, and distillation training, respectively. 
“Complexity” denotes the model computation complexity with respect to the number of tokens. 
“\#Data” denotes accumulated multimodal training data volume.
“\#Param” denotes the number of total parameters.
The best performance is highlighted in \textbf{\textcolor{red}{red}} and the second-best result is \textbf{\textcolor{blue}{blue}}.
}
\label{tab:resulits_llava}
\scalebox{1.0}
{
\begin{tabular}{lccccccccc}
\toprule
\multicolumn{1}{c}{Methods}              & Complexity                       & \#Data                      & \#Param                     & ai2d                                 & POPE                                 & MME-Perception                         & MMB-EN                               & SEED-Image                           & MMMU-Val                             \\ \midrule
{\color[HTML]{9B9B9B} MiniGPT-4}         & {\color[HTML]{9B9B9B} Quadratic} & {\color[HTML]{9B9B9B} -}    & {\color[HTML]{9B9B9B} 13B}  & {\color[HTML]{9B9B9B} -}             & {\color[HTML]{9B9B9B} -}             & {\color[HTML]{9B9B9B} 581.7}           & {\color[HTML]{9B9B9B} 23.0}          & {\color[HTML]{9B9B9B} -}             & {\color[HTML]{9B9B9B} -}             \\
{\color[HTML]{9B9B9B} LLaVA-1.5-7B}      & {\color[HTML]{9B9B9B} Quadratic} & {\color[HTML]{9B9B9B} 1.2M} & {\color[HTML]{9B9B9B} 7B}   & {\color[HTML]{9B9B9B} -}             & {\color[HTML]{9B9B9B} 85.9}          & {\color[HTML]{9B9B9B} 1510.7}          & {\color[HTML]{9B9B9B} 64.3}          & {\color[HTML]{9B9B9B} 66.1}          & {\color[HTML]{9B9B9B} }              \\
{\color[HTML]{9B9B9B} LLaVA-Phi}         & {\color[HTML]{9B9B9B} Quadratic} & {\color[HTML]{9B9B9B} 1.2M} & {\color[HTML]{9B9B9B} 2.7B} & {\color[HTML]{9B9B9B} -}             & {\color[HTML]{9B9B9B} 85.0}          & {\color[HTML]{9B9B9B} 1335.1}          & {\color[HTML]{9B9B9B} 59.8}          & {\color[HTML]{9B9B9B} -}             & {\color[HTML]{9B9B9B} -}             \\
{\color[HTML]{9B9B9B} Qwen-VL-Chat}      & {\color[HTML]{9B9B9B} Quadratic} & {\color[HTML]{9B9B9B} 1.4B} & {\color[HTML]{9B9B9B} 7B}   & {\color[HTML]{9B9B9B} -}             & {\color[HTML]{9B9B9B} -}             & {\color[HTML]{9B9B9B} 1487.5}          & {\color[HTML]{9B9B9B} 60.6}          & {\color[HTML]{9B9B9B} -}             & {\color[HTML]{9B9B9B} -}             \\
{\color[HTML]{9B9B9B} MobileVLM-3B}      & {\color[HTML]{9B9B9B} Quadratic} & {\color[HTML]{9B9B9B} 1.2M} & {\color[HTML]{9B9B9B} 3B}   & {\color[HTML]{9B9B9B} -}             & {\color[HTML]{9B9B9B} 84.9}          & {\color[HTML]{9B9B9B} 1288.9}          & {\color[HTML]{9B9B9B} 59.6}          & {\color[HTML]{9B9B9B} -}             & {\color[HTML]{9B9B9B} -}             \\
{\color[HTML]{9B9B9B} LLaVA-LLaMA3.2-1B} & {\color[HTML]{9B9B9B} Quadratic} & {\color[HTML]{9B9B9B} 1.2M} & {\color[HTML]{9B9B9B} 1B}   & {\color[HTML]{9B9B9B} 45.8}          & {\color[HTML]{9B9B9B} 86.9}          & {\color[HTML]{9B9B9B} 1229.2}          & {\color[HTML]{9B9B9B} 58.2}          & {\color[HTML]{9B9B9B} 59.2}          & {\color[HTML]{9B9B9B} 33.2}          \\
{\color[HTML]{9B9B9B} LLaVA-LLaMA3.2-3B} & {\color[HTML]{9B9B9B} Quadratic} & {\color[HTML]{9B9B9B} 1.2M} & {\color[HTML]{9B9B9B} 3B}   & {\color[HTML]{9B9B9B} 56.3}          & {\color[HTML]{9B9B9B} 86.5}          & {\color[HTML]{9B9B9B} 1366.0}          & {\color[HTML]{9B9B9B} 69.6}          & {\color[HTML]{9B9B9B} 67.3}          & {\color[HTML]{9B9B9B} 37.3}          \\ \midrule
VisualRWKV                               & Linear                           & 1.2M                        & 3B                          & -                                    & 83.1                                 & 1369.2                                 & {\color[HTML]{3166FF} \textbf{59.5}} & -                                    & -                                    \\
VL-Mamba                                 & Linear                           & -                           & 3B                          & -                                    & 84.4                                 & 1369.6                                 & 57.0                                 & -                                    & -                                    \\
Cobra                                    & Linear                           & 1.2M                        & 3.5B                        & -                                    & {\color[HTML]{FE0000} \textbf{88.4}} & -                                      & -                                    & -                                    &                                      \\
mmMamba-Linear                           & Linear                           & 1.7M                        & 2.7B                        & -                                    & 85.2                                 & 1303.5                                 & 57.2                                 & {\color[HTML]{3166FF} \textbf{62.9}} & 30.7                                 \\
mmMamba-Hybrid                           & Hybrid                           & 1.7M                        & 2.7B                        & -                                    & 86.7                                 & {\color[HTML]{3166FF} \textbf{1371.1}} & {\color[HTML]{FE0000} \textbf{63.7}} & {\color[HTML]{FE0000} \textbf{66.3}} & 32.3                                 \\ \midrule
% TransMamba-L                             & Linear                           & 0.6M                        & 0.5B                        & 26.2                                 & 64.6                                 & 550.1                                  & -                                    & -                                    & -                                    \\
TransMamba-H                             & Hybrid                           & 0.6M                        & 0.9B                        & {\color[HTML]{3166FF} \textbf{45.5}} & {\color[HTML]{3166FF} \textbf{87.0}} & 1270.0                                 & 56.6                                 & 59.2                                 & {\color[HTML]{3166FF} \textbf{33.3}} \\
TransMamba-L                             & Linear                           & 0.6M                        & 1.5B                        & 21.6                                 & 65.7                                 & 804.6                                  & -                                    & -                                    & 24.3                                 \\
TransMamba-H                             & Hybrid                           & 0.6M                        & 2.6B                        & {\color[HTML]{FE0000} \textbf{50.4}} & 84.2                                 & {\color[HTML]{FE0000} \textbf{1389.0}} & 58.6                                 & 61.5                                 & {\color[HTML]{FE0000} \textbf{34.1}} \\ \bottomrule
\end{tabular}
}
\vspace{-0.4cm}
\end{table*}

\begin{table}[]
\centering
\caption{
{\bf The accuracy of TransMamba on image classification (Train on $75\%$ data).} Acc1 represents the top-1 accuracy (\%), Acc5 represents the top-5 accuracy (\%). 
Results for the original Mamba are shown in \textbf{\textcolor{gray}{gray}}, while TransMamba results are presented in \textbf{bold}.
}
\label{tab:results_imageclassification}
\scalebox{1.0}
{
\begin{tabular}{cccc}
\toprule
                        & CIFAR & ImageNet$^S$             & ImageNet \\
\multirow{-2}{*}{Model} & Acc1  & Acc1                        & Acc1     \\ \midrule
DeiT-pretrain-T         & 88.2  & 87.6                        & 74.5     \\
DeiT-pretrain-S         & 88.2  & 87.8                        & 81.2     \\
DeiT-pretrain-B         & 88.4  & 88.1                        & 83.4     \\ \midrule
{\color[HTML]{808080}PMamba-T}                & {\color[HTML]{808080}78.6}  & {\color[HTML]{808080}84.5}                        & {\color[HTML]{808080}68.3}     \\
{\color[HTML]{808080}PMamba-S}                & {\color[HTML]{808080}81.2}  & {\color[HTML]{808080}85.8}                        & {\color[HTML]{808080}79.8}     \\
{\color[HTML]{808080}PMamba-B}                & {\color[HTML]{808080}82.3}  & {\color[HTML]{808080}86.6}                        & {\color[HTML]{808080}80.1}     \\
TransPMamba-T           & \textbf{83.9} {\color[HTML]{FE0000} ($\uparrow$5.3)}   & \textbf{86.6} {\color[HTML]{FE0000} ($\uparrow$2.1)}    & \textbf{70.8}  {\color[HTML]{FE0000} ($\uparrow$2.5)}    \\
TransPMamba-S           & \textbf{84.3} {\color[HTML]{FE0000} ($\uparrow$3.1)}  & \textbf{87.2}  {\color[HTML]{FE0000} ($\uparrow$1.4)}    & \textbf{80.8}  {\color[HTML]{FE0000} ($\uparrow$1.0)}    \\
TransPMamba-B           & \textbf{84.5} {\color[HTML]{FE0000} ($\uparrow$2.2)}  & \textbf{88.3}   {\color[HTML]{FE0000} ($\uparrow$1.7)}   & \textbf{81.3}   {\color[HTML]{FE0000} ($\uparrow$1.2)}   \\ \midrule
{\color[HTML]{808080}VMamba}                  & {\color[HTML]{808080}82.9}  & {\color[HTML]{808080}86.1}                        & {\color[HTML]{808080}82.2}     \\
TransVMamba             & \textbf{83.8} {\color[HTML]{FE0000} ($\uparrow$0.9)}  & \textbf{89.3} {\color[HTML]{FE0000} ($\uparrow$3.2)}     & \textbf{82.9}   {\color[HTML]{FE0000} ($\uparrow$0.7)}   \\ \midrule
{\color[HTML]{808080}ViM-T}                   & {\color[HTML]{808080}70.0}  & {\color[HTML]{808080}76.1}                        & -        \\
{\color[HTML]{808080}ViM-S}                   & {\color[HTML]{808080}73.9}  & {\color[HTML]{808080}84.5}                        & {\color[HTML]{808080}80.5}     \\
TransViM-T              & \textbf{76.6} {\color[HTML]{FE0000} ($\uparrow$6.6)}  & \textbf{81.2}   {\color[HTML]{FE0000} ($\uparrow$5.1)}                      & -        \\
TransViM-S              & \textbf{78.4}  {\color[HTML]{FE0000} ($\uparrow$4.5)}  & \textbf{85.2}  {\color[HTML]{FE0000} ($\uparrow$0.7)}        & \textbf{81.0}  {\color[HTML]{FE0000} ($\uparrow$0.5)}    \\ \bottomrule
\end{tabular}
}
\vspace{-0.4cm}
\end{table}
% \input{table_figs/tabBackboneVideo}
% \input{table_figs/tabReasoning}
%-----------------------------------------------------------------------------------------------

\subsection{Implementation Details.} \label{sec:4.2}
\noindent  \textbf{- Details of image classification}
We build our codebase following VMamba ~\citep{DBLP:journals/corr/abs-2401-10166}, PlainMamba ~\citep{DBLP:journals/corr/abs-2403-17695} and ViM ~\citep{DBLP:journals/corr/abs-2401-09417}. 
All Mamba variants are trained for 300 epochs using the AdamW optimizer ~\citep{DBLP:conf/iclr/LoshchilovH19} with a learning rate of 5e-4 and cosine learning rate scheduling. For VMamba, a batch size of 32 is used, while PMamba-T and PMamba-S use 128, and PMamba-B uses 64. Most experiments are conducted on a mix of 4 NVIDIA 3090 and 8 V100 GPUs.
The balancing hyperparameter $\alpha$ in Equation \ref{eq:loss_all} are set to 0.5.
Table ~\ref{tab:variants} summarizes the layer configurations, hidden dimensions, and parameter scales of various Mamba-based architectures for classification tasks. In contrast to the single PlainMamba configuration proposed in prior work ~\citep{DBLP:journals/corr/abs-2403-17695}, we introduce three scaled variants: PMamba-T, PMamba-S, and PMamba-B. For ViM, VMamba-T, and VideoMamba, we adopt the architectural settings described in ~\citep{DBLP:journals/corr/abs-2401-09417, DBLP:journals/corr/abs-2401-10166, DBLP:journals/corr/abs-2403-06977}.

\noindent  \textbf{- Details of visual question answering and multimodal reasoning}
We adopt the pre-trained CLIP-ViT-L/14 ~\citep{Radford2021LearningTV} as the vision encoder, followed by a two-layer MLP projector. Both teacher and student models are built upon the LLaMA-3.2~\citep{dubey2024llama} architecture (\eg, 3B/1B). The teacher employs the LLaVA training paradigm to obtain a LLaVA-LLaMA3.2-3B/1B model, while the student is constructed with reduced capacities based on the same architecture. Our Mamba model is then trained using 665K general captioning samples, with a batch size of 128, the Adam optimizer, and a learning rate of 2e-5. All models are trained for one epoch using 32 V100 GPUs.

\noindent  \textbf{- Details of video retrieval}
All Mamba models are trained for 5 epochs using the AdamW optimizer ~\citep{DBLP:conf/iclr/LoshchilovH19}, with an initial learning rate of 1e-4 and a cosine decay schedule. A frozen CLIP4Clip model serves as the teacher. Training is conducted with a batch size of 128 on 4 A100 GPUs.

\subsection{Main Results}\label{sec:4.3}

\noindent {\bf - Image Classfication.}  
%%%%% V 2
Table ~\ref{tab:results_imageclassification} presents classification results on CIFAR-100, ImageNet$^S$, and ImageNet1K. DeiT-Pretrain refers to models fine-tuned on CIFAR or ImageNet$^S$ after pre-training on ImageNet-1K ~\cite{DBLP:journals/ijcv/RussakovskyDSKS15}. Notably, PMamba, VMamba, and ViM are trained using the full training set, while their Trans- counterparts (TransPMamba, TransVMamba, and TransViM) are trained with only 75\% of the data. Additional results under varying data ratios are included in the ablation section.
Despite reduced training data, the TransMamba variants consistently outperform their Mamba-based baselines. Specifically, TransMamba-T surpasses PMamba-T by 2.1\%, TransVMamba exceeds VMamba by 3.2\%, and TransViM-T improves over ViM-T by 5.1\% on ImageNet$^S$. Meanwhile, both TransMamba-B and TransVMamba even outperform their respective teacher models, with TransVMamba achieving a 1.2\% gain.
Moreover, performance improves with model scale, suggesting that larger architectures benefit more from the transfer process. These results collectively demonstrate that knowledge distilled from Vision Transformers can be effectively transferred to Mamba architectures, leading to improved performance even under limited data regimes.

\begin{table}[t]
\centering
\caption{ {\bf
Comparison with different models on text-video retrieval datasets. }
Results for the original Mamba are shown in \textbf{\textcolor{gray}{gray}}, while TransMamba results are presented in \textbf{bold}.
}
\label{tab:results_video}
\scalebox{0.86}
{
\begin{tabular}{c|cccccc}
\toprule
Model & \multicolumn{3}{c}{MSR-VTT} & \multicolumn{3}{c}{DiDeMo} \\ 
                       & R@1$\uparrow$       & R@5$\uparrow$       & Mean R$\downarrow$  & R@1$\uparrow$      & R@5$\uparrow$       & Mean R$\downarrow$   \\ \midrule
CLIP4CLIP              & 42.1    & 71.9      & 15.7  & 42.3    & 69.1     & 18.6  \\
{\color[HTML]{808080}VideoMamba}             & {\color[HTML]{808080}40.9}    & {\color[HTML]{808080}69.2}    & {\color[HTML]{808080}16.4}  & {\color[HTML]{808080}40.8}    & {\color[HTML]{808080}68.8}    & {\color[HTML]{808080} 18.2}   \\
% VideoMamba-distill     & 40.8    & 69.0    & 79.1    & 39.3    & 67.0    & 78.3   \\
TransVideoMamba    & \textbf{41.6}    & \textbf{69.8}     & \textbf{16.1}   & \textbf{41.1}    & \textbf{68.8}    & \textbf{18.6} \\ \midrule
\end{tabular}}
\vspace{-0.4cm}
\end{table}

% \usepackage{multirow}
\begin{table}[]
\centering
\caption{
{\bf Comparison with different models on multimodal reasoning.}
“\#Param” denotes the number of total parameters.
% The best performance is highlighted in \textbf{\textcolor{red}{red}} and the second-best result is \textbf{\textcolor{blue}{blue}}.
}
\label{tab:multi-reasoning}
\scalebox{1.0}
{
\begin{tabular}{lccc}
\toprule
Model                          & \# Param & MathVista & MMMU \\ \midrule
Base Model                     &          &          &      \\
Llama-3.2-3B-Instruct-base     & 3B       &  -         &   37.3   \\
Llama-3.2-3B-Instruct-finetune & 3B       & 16.9      &   38.4   \\ \midrule
Our Model                      &          &           &      \\
TransMamba-Hybrid              & 2.6B     & 16.8      &  38.1    \\ \bottomrule
\end{tabular}
}
\vspace{-0.6cm}
\end{table}

%-----------------------------------------------------------------------------------------------

\noindent {\bf - Visual Question Answering. }
% Table ~\ref{tab:resulits_llava} reports results on the Visual Question Answering (VQA) task. Despite being trained on a smaller dataset, our proposed Trans-Hybrid-2.6B model achieves competitive performance, outperforming larger models on several metrics. Performance consistently improves with model scale, underscoring the efficacy of both the hybrid architecture and the distillation strategy.
% Compared to the pure Trans-Mamba variant, the Trans-Hybrid design yields consistently better results, highlighting the advantage of incorporating Transformer-derived knowledge into Mamba-based architectures. Importantly, our method achieves performance comparable to the teacher model while using significantly fewer parameters, demonstrating the efficiency of the knowledge transfer.
% Specifically, our method achieves state-of-the-art performance across multiple benchmarks, even when compared to models with significantly larger parameter counts. For instance, it outperforms mmMamba-Hybrid-2.7B by $17.9$ points on the MME-Perception benchmark and surpasses it by $1.8$ points on MMMU-val. Furthermore, our approach consistently demonstrates superior results compared to existing methods based on RWKV and Mamba architectures. Notably, our distillation strategy achieves these improvements using substantially less training data than prior art.
Table~\ref{tab:resulits_llava} reports results on the Visual Question Answering (VQA) task. Despite being trained on a significantly smaller dataset, our proposed Trans-Hybrid-2.6B model achieves competitive performance and outperforms larger models on multiple benchmarks. Performance improves consistently with model scale, underscoring the efficacy of the hybrid architecture and the knowledge distillation strategy. Compared to the pure Trans-Mamba variant, Trans-Hybrid yields superior results, demonstrating the advantage of integrating Transformer-derived knowledge. Notably, our method matches the performance of the teacher model with substantially fewer parameters, attesting to the efficiency of the knowledge transfer. In particular, it achieves state-of-the-art results—surpassing mmMamba-Hybrid-2.7B by 17.9 points on MME-Perception and by 1.8 points on MMMU-val—and consistently outperforms existing RWKV- and Mamba-based approaches, all while using considerably less training data than prior methods.
However, we observe that training large Mamba-based models is sensitive to initialization; poor initialization can lead to unstable gradients and convergence issues. Even with carefully designed initialization schemes, training remains less stable than Transformer-based counterparts—a challenge also noted in prior work. This suggests that initialization and training dynamics play a crucial role in scaling Mamba within multi-modal frameworks.
Overall, the results in Table ~\ref{tab:resulits_llava} validate the effectiveness of the proposed method, showing that Transformer knowledge can be successfully distilled into Mamba-based models while preserving strong performance and reducing model complexity.
%%%%结果出来后需要添加上
% \xw{Remember to add experiments details of GQA, VQA, VisWiz...}

\noindent {\bf - Text-Video Retrieval.}
Table ~\ref{tab:results_video} presents the results of TransMamba on the text-video retrieval task. Compared to VideoMamba, the proposed TransVideoMamba consistently achieves superior performance across all evaluation metrics. On the MSR-VTT dataset, for example, it improves R@1 by 0.5 percentage points and approaches the performance of CLIP4Clip, a strong Transformer-based teacher baseline. These findings further demonstrate the effectiveness of the proposed approach in video-centric multimodal settings.

\noindent {\bf - Multimodal Reasoning. }
Table ~\ref{tab:multi-reasoning} reports the results of TransMamba models on multimodal reasoning benchmarks. 
% The TransMamba-Hybrid variant consistently outperforms its teacher model across multiple metrics, despite being trained on a reduced dataset. Moreover, performance scales positively with model size, suggesting a synergistic effect between capacity and knowledge distillation. Under identical experimental conditions, the hybrid architecture outperforms its pure Mamba-based counterpart, highlighting the benefits of incorporating Transformer-derived knowledge. 
After distillation, the performance of TransMamba closely approaches that of the teacher model, indicating that knowledge within Transformers can be effectively transferred to Mamba architectures for multimodal reasoning tasks.
These results underscore the viability of transferring pre-trained Transformer knowledge into Mamba-based models, offering a promising pathway for developing efficient architectures that preserve strong multimodal reasoning capabilities.
% %%%%结果出来后需要添加上
% \xw{Remember to add experiments details of MathVista, MathVerse...}

\subsection{Ablation Studies}\label{sec:4.4}
We conduct a series of additional analyses to further examine the generalization and underlying properties of the proposed method. Specifically, we perform ablation studies on distillation strategies, varying training data proportions, and the selection of teacher model layers, with a focus on the impact of distilling from different depths within the final stage. We also extend our investigation to the RWKV architecture, demonstrating the broader applicability of our approach. All experiments are carried out on image classification benchmarks.
Furthermore, in the multimodal setting, we analyze the effect of varying the attention-to-Mamba ratio and their positional configurations within the Trans-Hybrid model. These studies provide insights into architectural trade-offs between reasoning capacity and computational efficiency, highlighting optimal configurations for enhanced multimodal performance.

% \begin{figure*}[]
% \centering
%   \begin{subfigure}{0.32\linewidth}
%     \includegraphics[width=\linewidth]{figs/results_acc_imnt_base.pdf}
%     \caption{PMamba / Accuracy}
%     \label{fig:Pmamba-acc}
%   \end{subfigure}
%   \hfill
%   \begin{subfigure}{0.31\linewidth}
%     \includegraphics[width=\linewidth]{figs/results_acc_imnt_base_vmamba.pdf}
%     \caption{VMamba / Accuracy}
%     \label{fig:Vmamba-acc}
%   \end{subfigure}
%   \hfill
%   \begin{subfigure}{0.33\linewidth}
%     \includegraphics[width=\linewidth]{figs/results_cifar_loss.pdf}
%     \caption{PMamba / Loss}
%     \label{fig:Pmamba-loss}
%   \end{subfigure}
% \caption{
% \textbf{Accuracy and loss curves under different architectures on image classification (\eg, PMamba and VMamba).}
% (a)(b) depict the accuracy cures of PMamba and VMamba on ImageNet$^S$, highlighting the substantially accelerated convergence of the proposed TransMamba (\eg, TransPMamba-B achieving up to \textbf{2.7× speed-up}) relative to baseline Mamba variants, while also \textbf{surpassing both the original Mamba and the teacher model (\eg, DeiT-Pretrain) in final accuracy}. (c) presents the loss curve of PMamba, further confirming the efficiency of TransMamba in optimization dynamics.
% }
% \label{fig:accuracy}
% \end{figure*}

\begin{figure*}[]
\centering
  \begin{subfigure}{0.24\linewidth}
    \includegraphics[width=\linewidth]{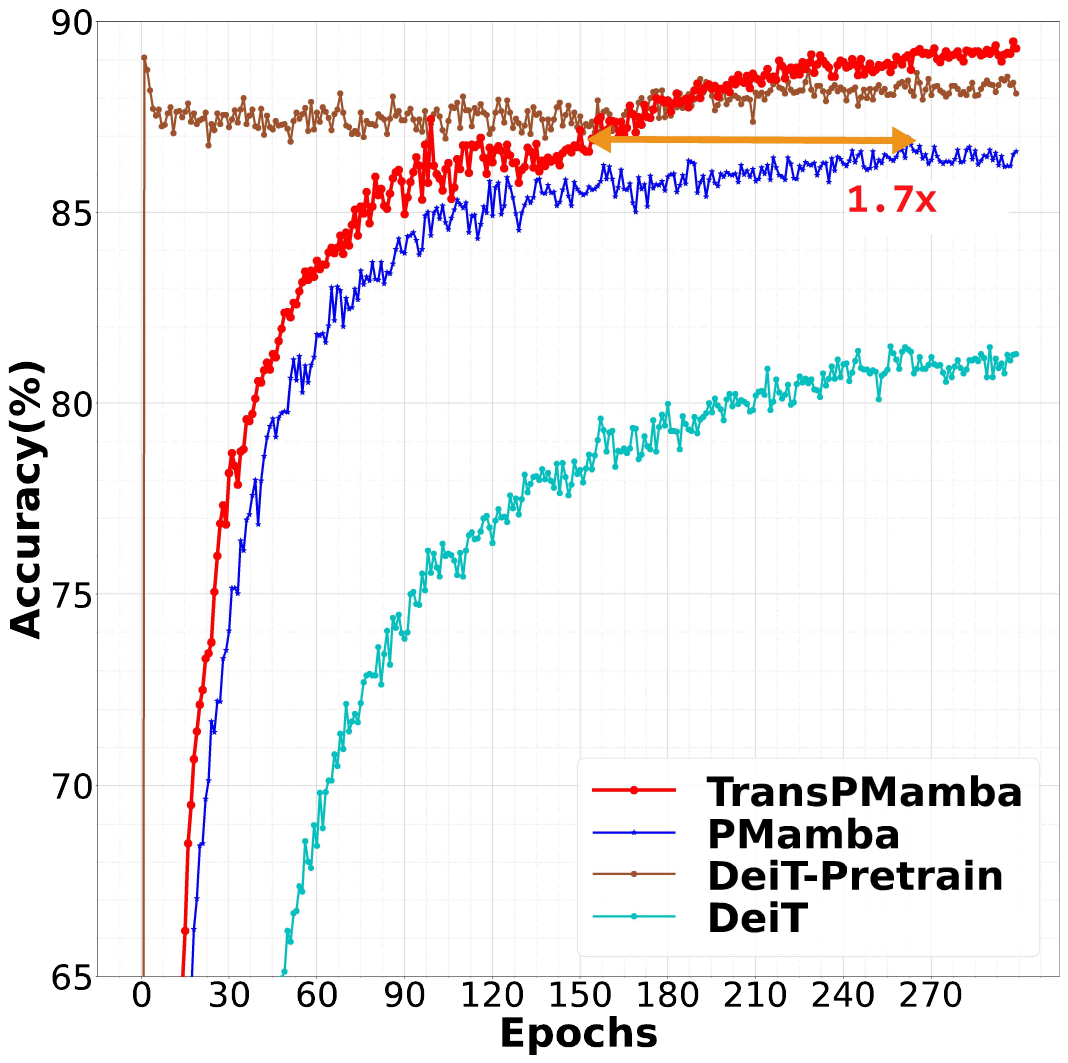}
    \caption{PMamba / Accuracy}
    \label{fig:Pmamba-acc}
  \end{subfigure}
  \hfill
  \begin{subfigure}{0.24\linewidth}
    \includegraphics[width=\linewidth]{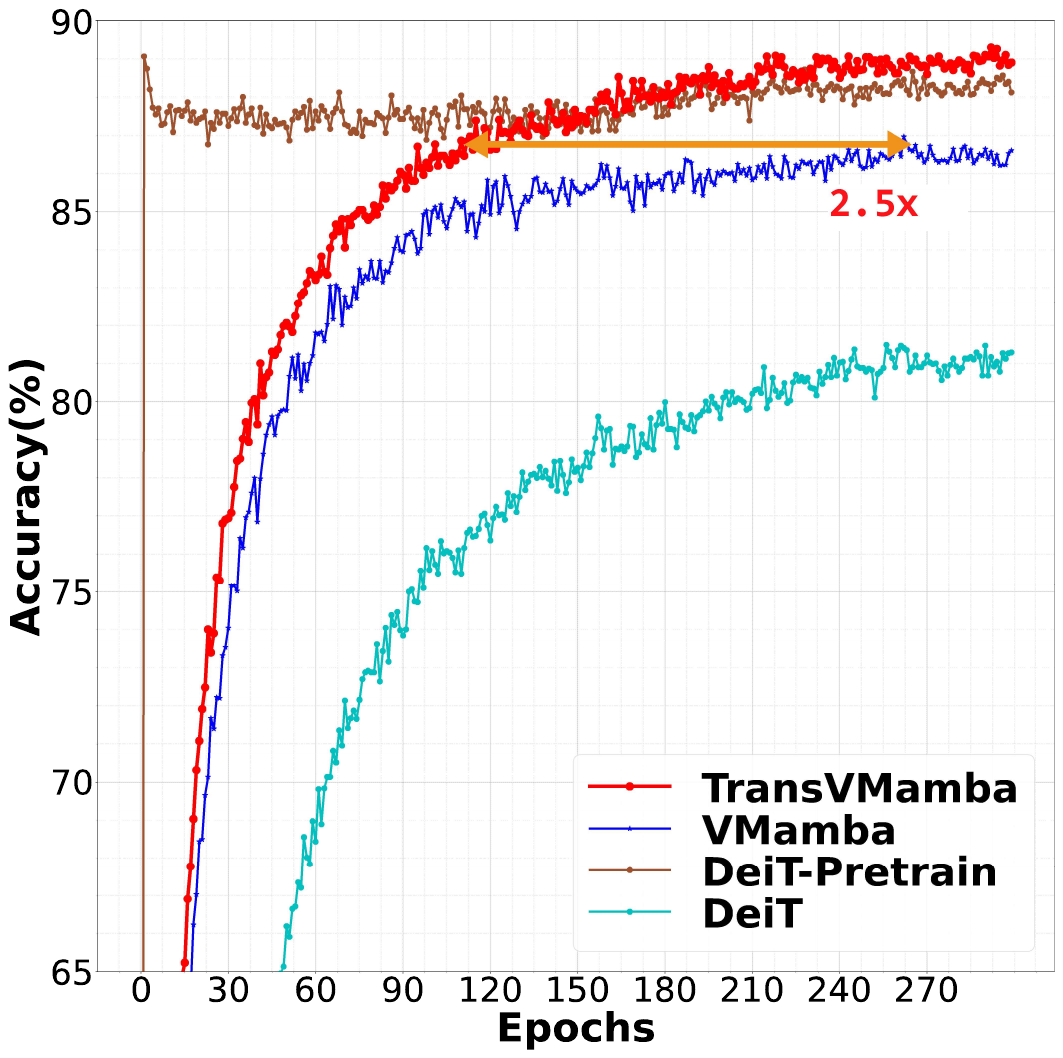}
    \caption{VMamba / Accuracy}
    \label{fig:Vmamba-acc}
  \end{subfigure}
  \hfill
  \begin{subfigure}{0.24\linewidth}
    \includegraphics[width=\linewidth]{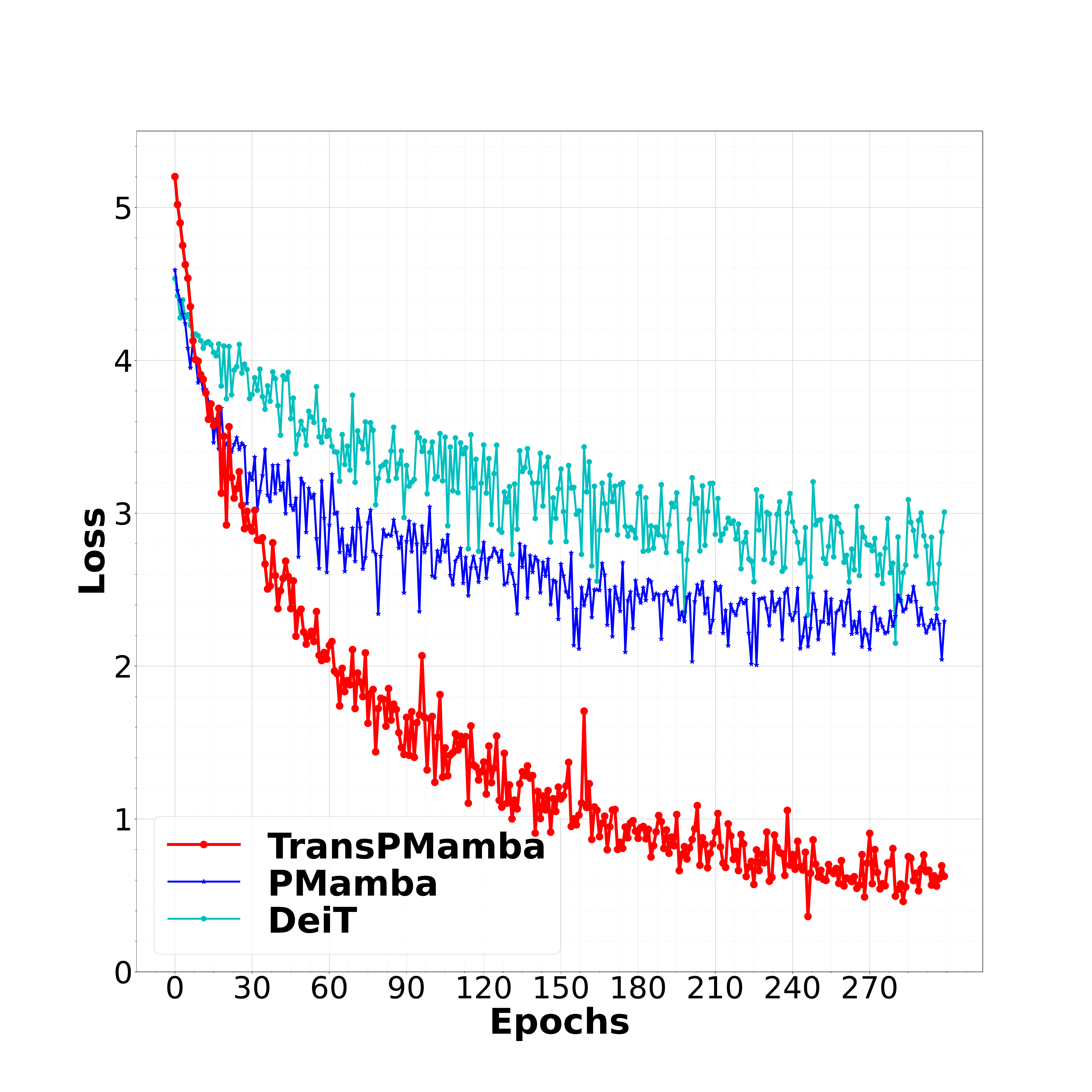}
    \caption{PMamba / Loss}
    \label{fig:Pmamba-loss1}
  \end{subfigure}
  \hfill
  \begin{subfigure}{0.24\linewidth}
    \includegraphics[width=\linewidth]{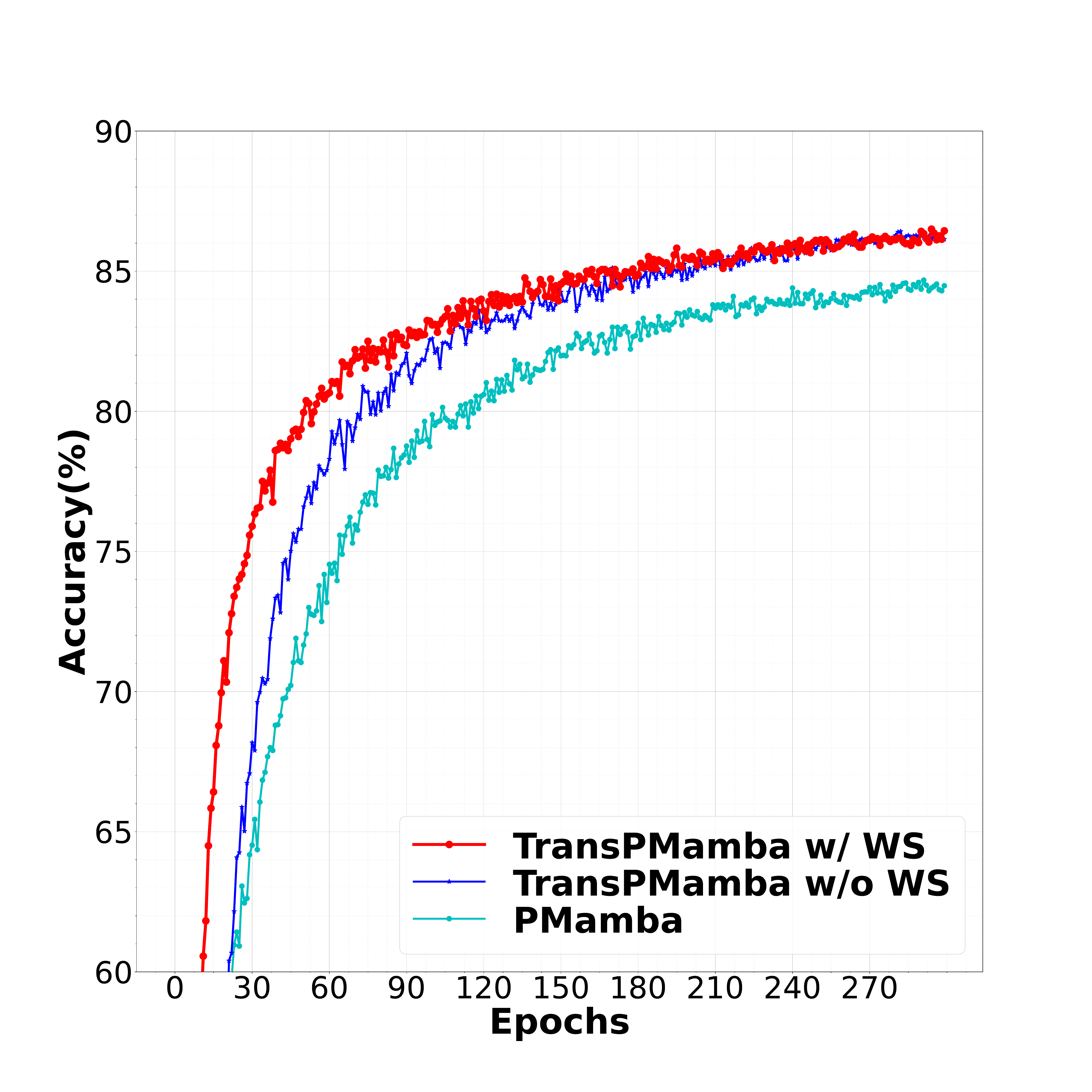}
    \caption{PMamba / Loss}
    \label{fig:Pmamba-loss2}
  \end{subfigure}
\caption{
\textbf{Accuracy and loss curves under different architectures on image classification (\eg, PMamba and VMamba).}
(a)(b) depict the accuracy cures of PMamba and VMamba on ImageNet$^S$, highlighting the substantially accelerated convergence of the proposed TransMamba (\eg, TransPMamba-B achieving up to \textbf{2.7× speed-up}) relative to baseline Mamba variants, while also \textbf{surpassing both the original Mamba and the teacher model (\eg, DeiT-Pretrain) in final accuracy}. (c) presents the loss curve of PMamba, further confirming the efficiency of TransMamba in optimization dynamics.
}
\label{fig:accuracy}
\vspace{-0.4cm}
\end{figure*}

% \usepackage{multirow}

\begin{table}[]
\centering
\caption{
{\bf Ablation studies on different distillation strategies for image classification (TransPMamba-T).}
Acc1 accuracy results on the CIFAR and ImageNet$^S$ datasets.
The best performance is highlighted in \textbf{bold}.
}
\label{tab:aba_distill}
\scalebox{1.0}
{
% \begin{tabular}{c|cc}
% \midrule
% \multirow{2}{*}{Distill Strategy} & \multicolumn{2}{c}{ImageNet-Subset} \\
%                        & \multicolumn{1}{c|}{Acc1}   & Acc5  \\ \midrule
% feature distill          &   78.56   &    93.72       \\
% logits distill           &  84.36    &  97.12         \\
% WSAB &   87.10    &  97.22        \\ \midrule
% \end{tabular}
\begin{tabular}{lcc}
\toprule
\multicolumn{1}{c}{Distill Strategy}                             & CIFAR & ImageNet$^S$ \\ \midrule
Baseline                                                       & 78.6  & 84.5            \\
Logits                                                        & 82.5  & 84.4            \\
Feature                                                      & 81.5  & 78.6            \\
Logits+Feature                                               & 82.6  & 85.2            \\
Layer+Feature+Logits                                         & 82.8  & 85.4            \\
Direction+Feature+Logits                                     & 83.7  & \textbf{86.6}            \\
Adapt-Layer+Feature+Logits                                 & 83.4  & 86.4            \\
Direction+Adapt-Layer+Feature+Logits                         & \textbf{83.9}  & \textbf{86.6}            \\ \bottomrule
\end{tabular}
}
\vspace{-0.2cm}
\end{table}

% \usepackage{multirow}
\begin{table}[]
\centering
\caption{
{\bf Ablation studies on different training data proportions for image classification (\eg, TransPMamba-S).}
Acc1/Acc5 accuracy results on the ImageNet$^S$ datasets.
The best performance is highlighted in\textbf{bold}.
}
\label{tab:aba_datasize}
\scalebox{1.2}
{
\begin{tabular}{cccc}
\toprule
\multirow{2}{*}{Model}         & \multirow{2}{*}{Data Proportions} & \multicolumn{2}{c}{ImageNet}  \\ \cline{3-4} 
                               &                                   & Acc1          & Acc5          \\ \midrule
\multirow{4}{*}{TransPMamba-S} & 25\%                              & 83.2          & 96.4          \\
                               & 50\%                              & 87.0          & \textbf{97.7} \\
                               & 75\%                              & \textbf{87.2} & \textit{97.2} \\
                               & 100\%                             & 87.0          & 97.2          \\ \bottomrule
\end{tabular}
}
\vspace{-0.6cm}
\end{table}

% \usepackage{multirow}
\begin{table}[t]
\centering
\caption{
{\bf Ablation studies on supervisory layer selection in transformer teachers for image classification (\eg, TransPMamba-S).}
Acc1/Acc5 accuracy results on the ImageNet$^S$ datasets.
The best performance is highlighted in\textbf{bold}.
}
\label{tab:aba_one}
\scalebox{1.2}
{
\begin{tabular}{cccc}
\toprule
\multirow{2}{*}{Model}         & \multirow{2}{*}{\makecell{Transformer Layer \\ Index}} & \multicolumn{2}{c}{ImageNet}  \\ \cline{3-4} 
                               &                                          & Acc1          & Acc5          \\ \midrule
\multirow{3}{*}{TransPMamba-S} & shallow (4)                              & 84.3          & 96.9          \\
                               & middle (8)                               & 83.8          & 96.3          \\
                               & final (12)                               & \textbf{87.2} & \textbf{97.2} \\ \bottomrule
\end{tabular}
}
\vspace{-0.2cm}
\end{table}
\noindent\textbf{- Distillation strategy.} 
Table ~\ref{tab:aba_distill} investigates the impact of various distillation strategies. Conventional techniques (\eg, logit-based and feature-level matching) often lead to suboptimal or even degraded performance when applied across architectures, particularly from Transformer teachers to Mamba-based students. This underscores the difficulty of cross-architecture knowledge transfer and suggests that standard distillation methods are insufficient in this context.
In contrast, our proposed multi-directional distillation framework yields a 2.1\% point improvement on ImageNet over the non-distilled baseline, while its adaptive variant achieves a 1.9\% point gain. The full TransMamba configuration further improves performance by 2.1\% point, highlighting the effectiveness of the proposed strategy in facilitating robust knowledge transfer to Mamba architectures.

\noindent\textbf{- Varying training data proportions.}
Table \ref{tab:aba_datasize} examines the performance of TransMamba under varying training data scales. Remarkably, the model achieves state-of-the-art results using only around 50\% of the full dataset, with additional data yielding diminishing returns. This highlights the data efficiency of the proposed approach and demonstrates its scalability across different data regimes.

\noindent\textbf{- The selection of teacher model layers.}
Table ~\ref{tab:aba_one} investigates the effect of selecting different Transformer layers as supervision sources for distillation. The results reveal that using features from shallow layers adversely impacts the performance of the Mamba-based student, while supervision from the final Transformer layer yields notable improvements. This indicates that deeper layers encapsulate richer semantic and task-specific information, making them more effective for cross-architecture knowledge transfer.

% \input{table_figs/TabAbaRWKV}
% \noindent\textbf{Generalization to Transformer architecture.}
% Table \ref{tab:aba_archi} reports the performance of applying the proposed method to the RWKV architecture using the adaptive distillation strategy. TransRWKV consistently outperforms the baseline across all metrics, with TransRWKV-T achieving a 1.1\% gain in Acc@1 on ImageNet-S. These findings highlight the generalizability of our approach beyond Mamba-based models, demonstrating that the core components of the framework are transferable to structurally distinct architectures, thereby underscoring its broader applicability.

% \usepackage{multirow}
\begin{table}[]
\centering
\caption{
{\bf Ablation studies on different architectures for image classification (\eg, RWKV).}
Acc1/Acc5 accuracy results on the ImageNet$^S$ datasets.
Results for the original RWKV are shown in \textbf{\textcolor{gray}{gray}}, while TransRWKV results are presented in \textbf{bold}.
}
\label{tab:aba_archi}
\scalebox{1.2}
{
\begin{tabular}{c|ll}
\toprule
\multirow{2}{*}{Model} & \multicolumn{2}{c}{ImageNet$^S$} \\
                       & \multicolumn{1}{c|}{Acc1}   & Acc5  \\ \midrule

{\color[HTML]{808080}RWKV-T}                 &   {\color[HTML]{808080}83.1}                          &  {\color[HTML]{808080}95.3}     \\
{\color[HTML]{808080}RWKV-S}                 &     {\color[HTML]{808080}84.5}                        &   {\color[HTML]{808080}95.7}    \\
TransRWKV-T            &      \textbf{84.2} {\color[HTML]{FE0000} ($\uparrow$1.1)}                      &   \textbf{96.3} {\color[HTML]{FE0000} ($\uparrow$1.0)}   \\
TransRWKV-S            &      \textbf{85.2} {\color[HTML]{FE0000} ($\uparrow$0.7)}                      &   \textbf{96.5} {\color[HTML]{FE0000} ($\uparrow$0.8)}   \\ \midrule
\end{tabular}}
\vspace{-0.6cm}
\end{table}
\noindent\textbf{- Generalization to RWKV architecture.}
Table \ref{tab:aba_archi} reports the performance of applying the proposed method to the RWKV architecture using the adaptive distillation strategy. TransRWKV consistently outperforms the baseline across all metrics, with TransRWKV-T achieving a 1.1\% gain in Acc@1 on ImageNet-S. These findings highlight the generalizability of our approach beyond Mamba-based models, demonstrating that the core components of the framework are transferable to structurally distinct architectures, thereby underscoring its broader applicability.

% \usepackage{multirow}
\begin{table*}[]
\centering
\caption{
{\bf Ablation studies on varying the attention-to-Mamba ratio and their positional for MLLMs.}
“\#Data” denotes accumulated multimodal training data volume.
“\#Param” denotes the number of total parameters.
The best performance is highlighted in \textbf{\textcolor{red}{red}} and the second-best result is \textbf{\textcolor{blue}{blue}}.
}
\label{tab:ratio-position}
\scalebox{1.0}
{
\begin{tabular}{lccccc}
\toprule
Model                         & \#Data & \#Param & POPE                                 & MME-Perception                         & MMMU-Val                             \\ \midrule
Base Model                    &        &         &                                      &                                        &                                      \\
LLaVA-LLaMA-3.2-1B-Instruct   & 1.2M   & 1B      & {\color[HTML]{34CDF9} \textbf{86.9}} & 1229.2                                 & {\color[HTML]{34CDF9} \textbf{33.2}} \\ \midrule
Our Model                     &        &         &                                      &                                        &                                      \\
TransMamba-Hybrid-50\%-low    & 0.6M   & 0.7B    & 81.3                                 & 839.5                                  & -                                    \\
TransMamba-Hybrid-50\%-high   & 0.6M   & 0.7B    & 85.9                                 & 1080.2                                 & 28.2                                 \\ \midrule
TransMamba-Hybrid-13\%-high & 0.6M   & 0.9B    & {\color[HTML]{FE0000} \textbf{87.0}} & {\color[HTML]{34CDF9} \textbf{1270.0}} & {\color[HTML]{FE0000} \textbf{33.3}} \\
TransMamba-Hybrid-25\%-high   & 0.6M   & 0.9B    & {\color[HTML]{FE0000} \textbf{87.0}} & {\color[HTML]{FE0000} \textbf{1284.9}} & 32.1                                 \\
TransMamba-Hybrid-50\%-high   & 0.6M   & 0.7B    & 85.9                                 & 1080.2                                 & 28.2                                 \\
TransMamba-Hybrid-75\%-high   & 0.6M   & 0.6B    & 70.6                                 & 778.5                                  & 23.9                                 \\ \bottomrule
\end{tabular}
}
\vspace{-0.4cm}
\end{table*}

\noindent\textbf{- Varying the attention-to-Mamba ratio and their positional.}
Table \ref{tab:ratio-position} reports an ablation study on the ratio and arrangement of attention and Mamba blocks within the Trans-HybridMamba architecture under multimodal settings. Results show that increasing the proportion of attention blocks generally leads to improved performance, despite their higher parameter count relative to Mamba blocks. In contrast, the specific placement of Mamba blocks exerts limited influence on final outcomes, indicating that the architecture is relatively insensitive to block ordering and robust across different configurations.

\begin{figure*}[t]
  \centering
   \includegraphics[width=0.99\linewidth]{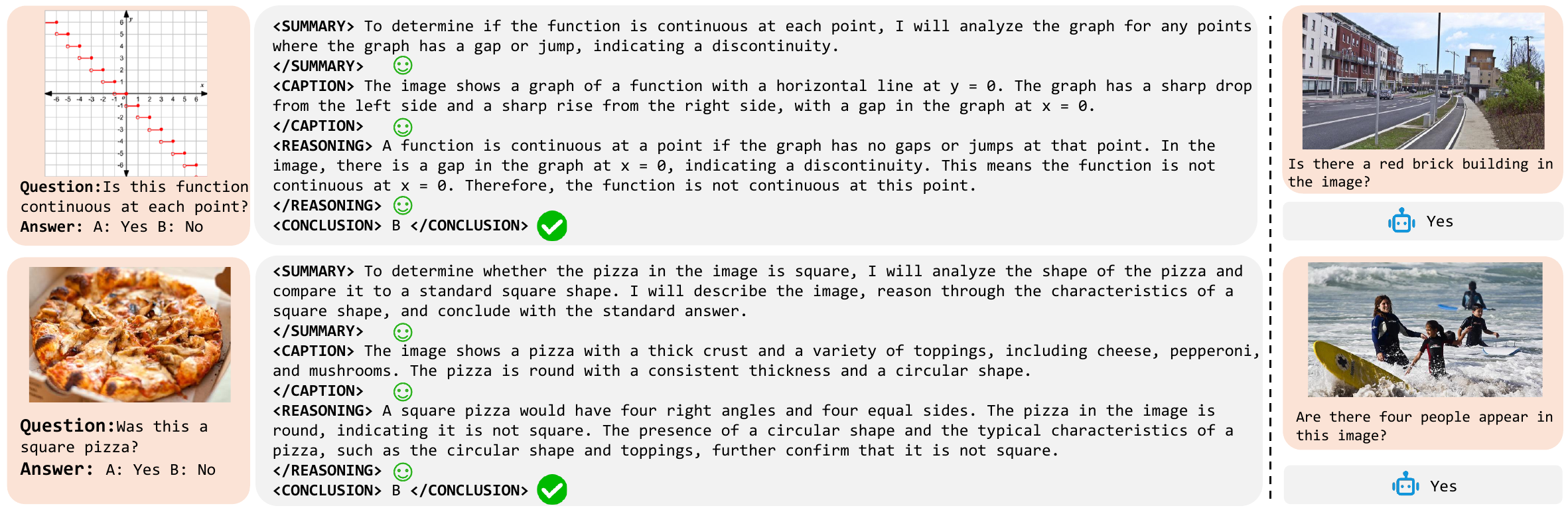}

   \caption{
   \textbf{Case studies on evaluation samples on multimodal reasoning (left) and visual question answering (right).}
   % The TransMamba model demonstrates strong understanding of query intent and provides accurate responses, while effectively retaining context across multiple turns.
   }
   \label{fig:transllava visual}
\vspace{-0.6cm}
\end{figure*}

\subsection{Visualization}\label{sec:4.5}
\noindent\textbf{- Visualization of accuracy/function loss on image classification.}
Figure \ref{fig:accuracy} illustrates the convergence dynamics of PMamba and VMamba by comparing their accuracy and loss curves. TransMamba not only converges substantially faster—achieving comparable performance in roughly one-third of the training time—but also surpasses the final accuracy of the baseline models, demonstrating both the training efficiency and effectiveness of the proposed method.
% In addition, Figure \ref{fig:accuracy} (d) highlights the impact of weight sub-cloning during initialization on TransPMamba. Models initialized with sub-cloned weights exhibit significantly faster convergence in the early training stages, underscoring the critical role of informed initialization in enhancing the optimization process for Mamba-based architectures.

% \noindent\textbf{Visualization of heatmap on image classification.}
% To validate the efficiency of our TransMamba, we present the loss function convergence graph and accuracy graph during the Mamba training process in Figure \ref{fig:accuracy}.
% It is evident from the results that TransMamba exhibits faster convergence rates and improved accuracy, confirming the training efficiency discussed in Section \ref{sec:intro}.  

\noindent\textbf{- Qualitative results of visual question answering.}
Figure \ref{fig:transllava visual} (right) presents qualitative visualizations of TransMamba on multimodal Visual Question Answering (VQA) tasks. The model demonstrates strong intent understanding and context-aware response generation. In particular, it effectively maintains dialogue coherence across multiple turns, benefiting from Mamba's long-sequence modeling capabilities. Moreover, TransMamba exhibits robust performance across diverse question types—including open-ended and multiple-choice formats—highlighting its versatility and generalization in complex multimodal reasoning scenarios.

\noindent\textbf{- Qualitative results of multimodal reasoning.}
Figure \ref{fig:transllava visual} (left) illustrates the reasoning trajectory of TransMamba on multimodal reasoning tasks. The model demonstrates a strong capacity to analyze input content and synthesize relevant information to generate well-justified conclusions. Additionally, it adheres closely to the predefined output formats of the dataset, highlighting its ability to produce structured and task-compliant responses. These qualitative results further underscore the effectiveness of TransMamba in handling complex multimodal reasoning scenarios.

\section{Conclusion}
\label{sec:conlu}

\noindent \textbf{- Conclusion.}
% This work proposes a two-stage knowledge transfer framework for adapting pre-trained Transformers to Mamba architectures. It addresses cross-architectural discrepancies via selective weight inheritance and sub-cloning, followed by adaptive multi-directional distillation to preserve multimodal reasoning. The method achieves strong performance across diverse tasks and Mamba variants with reduced data and computational cost, bridging the gap between Transformer capability and Mamba efficiency.
This research introduces a novel two-stage knowledge transfer framework designed to address three key challenges in adapting pre-trained Transformer models to the Mamba architecture: cross-architecture learning, multi-order dependency in selective scanning, and transfer of multimodal reasoning capabilities. The first stage involves constructing Mamba-based architectures by selectively inheriting key weights from pre-trained Transformers, excluding QKV projection layers, while employing a selective weight sub-cloning mechanism and layer-wise initialization strategy to mitigate architectural discrepancies. The second stage focuses on aligning output distributions between transformer and mamba architectures through an adaptive multi-directional distillation strategy that preserves hierarchical feature semantics.

The proposed framework demonstrates strong generalization across various tasks, including image classification, visual question answering, text-video retrieval, and multimodal reasoning, as well as diverse Mamba-based model families such as PlainMamba, VMamba, ViM, and VideoMamba. It achieves state-of-the-art performance compared to larger models and existing methods like RWKV and Mamba architectures. Additionally, the method uses less training data than previous approaches, showcasing its efficiency.

Key contributions include the development of a fast and universal transfer framework, a selective subcloning mechanism for effective parameter reuse, an adaptive multi-directional distillation method for feature alignment, and comprehensive validation across multiple backbones and downstream tasks. This work effectively bridges the gap between the inference strength of multimodal Transformer models and the architectural efficiency of Mamba, paving the way for scalable and efficient multimodal reasoning systems.

% \section*{Acknowledgments}
% This should be a simple paragraph before the References to thank those individuals and institutions who have supported your work on this article.

\bibliographystyle{IEEEtran}
\bibliography{main}

 

% \end{IEEEbiography}



\vfill

\end{document}